\documentclass[lettersize,journal]{IEEEtran}
\usepackage{times}

\usepackage[noadjust]{cite}

\usepackage{amsmath,amssymb,amsfonts,amsthm,dsfont,mathtools}
\usepackage{nicematrix}

\usepackage{adjustbox}
\usepackage{graphicx,tabularx,adjustbox,color}
\usepackage{multicol}
\usepackage{multirow}

\usepackage{algorithm,algorithmicx,listings}
\usepackage[noend]{algpseudocode}

\usepackage[font=small,labelfont=bf]{caption}
\usepackage[font=small,labelfont=bf]{subcaption}
\usepackage{diagbox}  

\usepackage{array}
\usepackage{textcomp}
\usepackage{stfloats}
\usepackage{verbatim}
\usepackage{dirtytalk}
\MakeRobust{\say}

\usepackage[breaklinks=true, colorlinks, bookmarks=true]{hyperref}

\DeclareMathOperator*{\argmin}{arg\,min}

\theoremstyle{definition}

\newtheorem{problem}{Problem}
\newtheorem*{problem*}{Problem}

\newcommand{\new}[1]{{\color{blue}{#1}}}  

\newcommand{\calB}{{\cal B}}

\newcommand{\calK}{{\cal K}}

\newcommand{\calM}{{\cal M}}

\newcommand{\calO}{{\cal O}}
\newcommand{\calP}{{\cal P}}
\newcommand{\calQ}{{\cal Q}}

\newcommand{\calS}{{\cal S}}

\newcommand{\calU}{{\cal U}}

\newcommand{\bfp}{\mathbf{p}}
\newcommand{\bfq}{\mathbf{q}}

\newcommand{\bfu}{\mathbf{u}}

\newcommand{\bftheta}{\boldsymbol{\theta}}

\newcommand{\bfxi}{\boldsymbol{\xi}}

\newcommand{\bfI}{\mathbf{I}}

\newcommand{\bfR}{\mathbf{R}}

\newcommand{\bfT}{\mathbf{T}}

\newcommand{\bbR}{\mathbb{R}}

\begin{document}

\title{Neural Configuration-Space Barriers for Manipulation Planning and Control}


\author{Kehan Long$^1$, Ki Myung Brian Lee$^1$, Nikola Raicevic$^1$, Niyas Attasseri$^1$, Melvin Leok$^1$, and Nikolay Atanasov$^1$%
\thanks{We gratefully acknowledge support from NSF CCF-2112665 (TILOS) and from the Ministry of Trade, Industry and Energy (MOTIE), Korea, under the Strategic Technology Development Program supervised by the Korea Institute for Advancement of Technology (KIAT) [Grant No. P0026052].}%
\thanks{$^1$The authors are with Contextual Robotics Institute, University of California San Diego, La Jolla, CA 92093, USA.}%
}



%

\maketitle

\begin{abstract}
Planning and control for high-dimensional robot manipulators in cluttered dynamic environments require computational efficiency and robust safety guarantees. Inspired by recent advances in learning configuration-space distance functions (CDFs) as representations of robot bodies, we propose a unified approach for motion planning and control that formulates safety constraints as CDF barriers. A CDF barrier approximates the local free configuration space, substantially reducing the number of collision-checking operations during motion planning. However, learning a CDF barrier with a neural network and relying on online sensor observations introduces uncertainties that must be considered during control synthesis. To address this, we develop a distributionally robust CDF barrier formulation for control that accounts for modeling errors and sensor noise without assuming a known underlying distribution. Simulations and hardware experiments on a UFactory xArm6 manipulator show that our neural CDF barrier formulation enables efficient planning and robust safe control in cluttered and dynamic environments, relying only on onboard point-cloud observations.

Note to Practitioners—Robotic manipulators operating in cluttered dynamic environments, such as warehouses or assembly lines, face challenges in ensuring safe and efficient real-time motion. Traditional methods often rely on detailed models of the environment and frequent collision checks, which are impractical when using onboard sensors and when obstacles move unpredictably. This paper introduces a new framework that integrates motion planning and control using learned configuration-space distance functions (CDFs) to identify local safe regions, or “bubbles.” These bubbles substantially reduce the number of collision checks during planning, reducing computation time. For real-time control, the method employs a robust safety filter that explicitly accounts for errors in the learned CDF model and sensor measurements, maintaining safe operation even with dynamic obstacles. Our approach is demonstrated on a UFactory xArm6 manipulator and can be extended to other robot manipulators operating in dynamic environments. We hope this approach improves productivity and safety in industrial settings where accurate environment models are unavailable. 
\end{abstract}

\section{Introduction}
Efficient planning and safe control for high-degree-of-freedom (DoF) robotic manipulators are a central challenge in robotics, particularly when operating in cluttered and dynamic environments with noisy sensor measurements. Sampling-based motion planners, such as rapidly-exploring random trees (RRT) \cite{lavalle2001rapidly, wei2018method} and probabilistic roadmaps (PRM) \cite{kavraki1996probabilistic}, have gained widespread adoption for manipulator motion planning because they effectively explore high-dimensional spaces without exhaustive configuration-space discretization. Despite their strengths, these methods rely on frequent collision checks, making the efficiency and accuracy of collision checking a critical bottleneck. Moreover, in highly dynamic settings with sensor noise and rapidly changing obstacles, planning alone is insufficient to ensure safety during execution, necessitating real-time safety-aware control strategies.

To address these challenges, signed distance functions (SDFs) have gained popularity for continuously encoding the distance from any point in the workspace to the boundary of obstacles, thereby enabling precise collision checks and facilitating gradient-based optimization. Recently, neural networks have been leveraged to learn SDF representations of both environments \cite{deepsdf, sitzmann2020implicit, Long_learningcbf_ral21} and robot bodies \cite{koptev_neural_jsdf_2022, li2024representing, zhu2024efficient}, providing rich shape modeling along with efficient distance and gradient queries. Building on these advancements, neural SDFs have been adopted in motion planning \cite{zhu2024efficient, quintero_2024_ICRA_sdf_planning,voxblox_Oleynikova} and optimization-based safe control \cite{Long_learningcbf_ral21, singletary_2023_ral_sdf_cbf_qp} for mobile robots, yielding promising results in complex, dynamic scenarios.

\begin{figure}[t]
\centering
\subcaptionbox{Work space\label{fig:1a}}{
\includegraphics[width=0.47\linewidth]{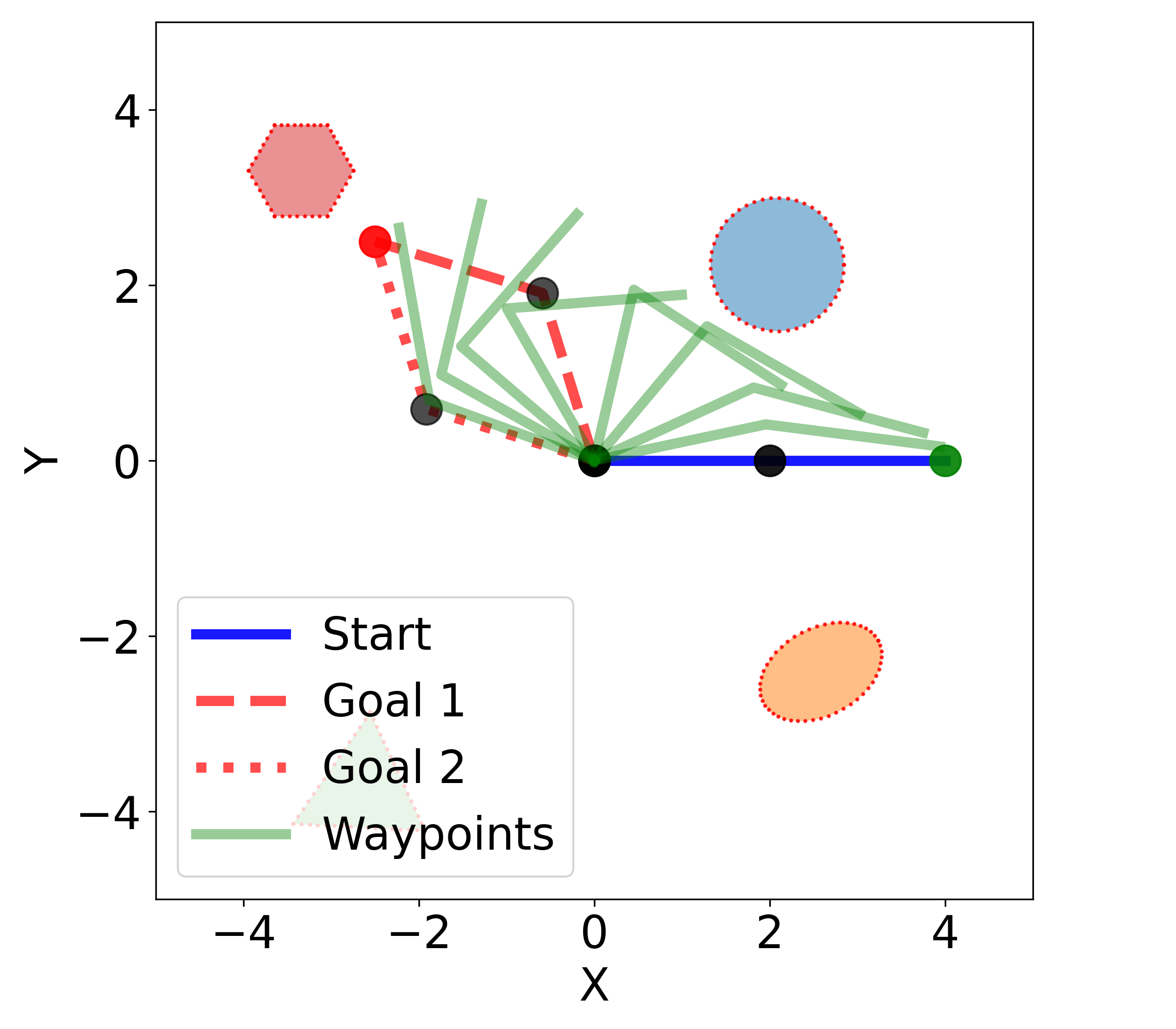}}%
\hfill%
\subcaptionbox{Configuration space\label{fig:1b}}{
\includegraphics[width=0.5\linewidth]{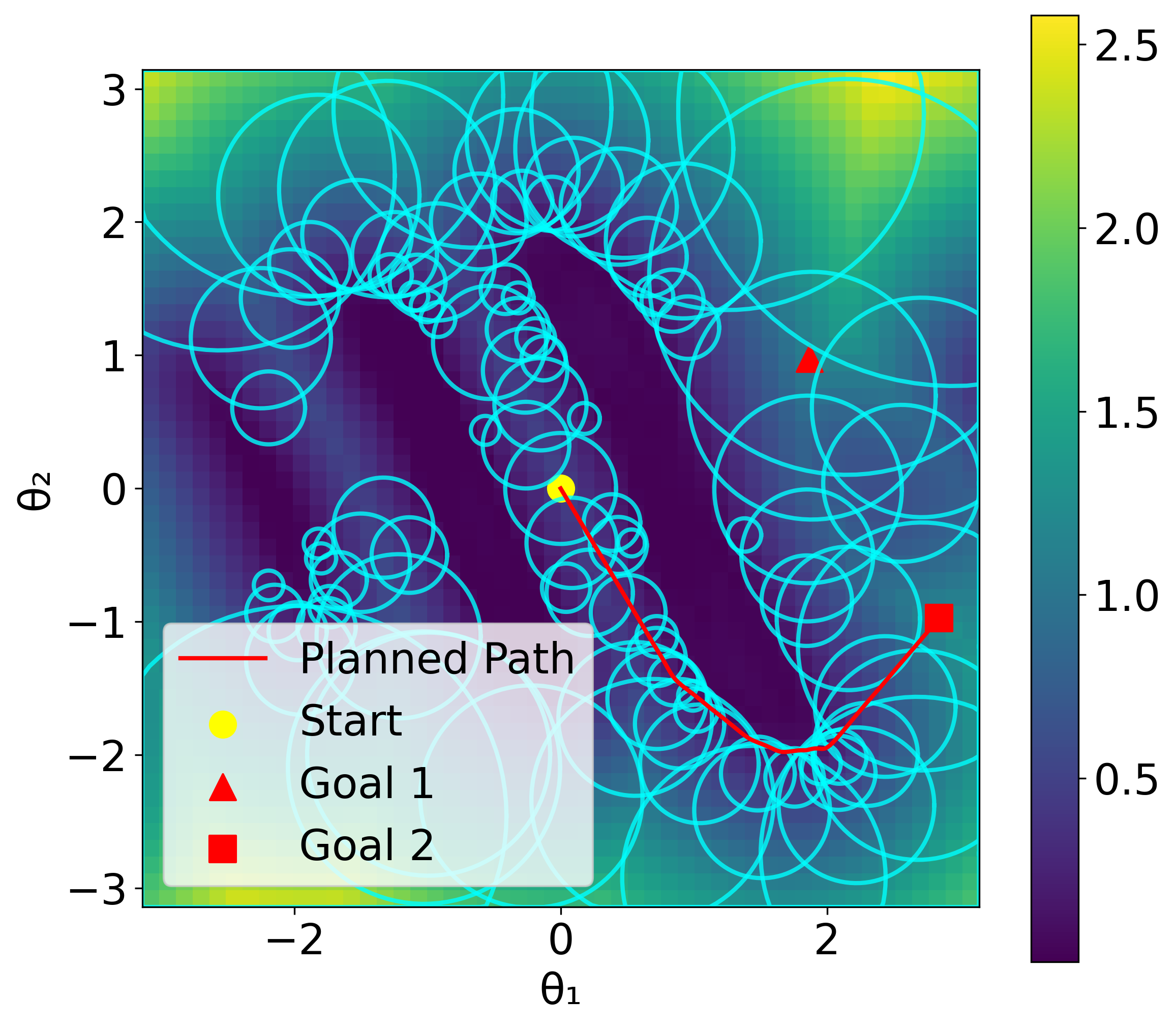}}
\caption{Illustration of sampling-based planning using configuration-space spheres to represent collision-free space for a planar 2-link robot arm.
(a) The robot’s initial configuration (solid blue lines) and two potential goal configurations (dashed red). Four obstacles are present, represented as point-cloud data. Intermediate waypoints from the planned trajectory appear in light green. 
(b) The 2D configuration space (\(\theta_1\) vs.\ \(\theta_2\)) of the 2-link robot, where the color map indicates the learned configuration-space distance function (CDF) value (darker regions denote proximity to collision). Cyan circles represent safe bubbles derived from the CDF, forming a graph of collision-free regions. A smooth, optimized trajectory (red) connects the start configuration (yellow circle) to the closest goal (red square).}
\label{fig:2d_example_illustrate}
\vspace{-3ex}
\end{figure}

To extend the benefits of SDF to high-DoF manipulators, recent work \cite{li2024cdf_rss} introduced a \emph{configuration-space distance function} (CDF) that directly represents distances in configuration space. In a CDF, the distance is measured in joint space between a given configuration and a workspace point, eliminating the need for repeated transformations between task and configuration spaces. Like neural SDFs, neural CDFs offer a compact and differentiable representation, supporting efficient gradient-based motion planning \cite{li2024cdf_rss} and reactive control \cite{chi2024safe}. However, as learning-based representations inevitably introduce modeling errors, real-world operations must also contend with sensor noise and other uncertainties.

Motivated by these challenges, we propose an approach that integrates a neural CDF barrier into both motion planning and control, addressing the twin requirements of computational efficiency and safety under uncertainty. The \textbf{contributions} of this work are summarized as follows.

\begin{itemize}
    \item We propose a unified neural configuration-space distance function that estimates distances to workspace obstacles and to self-collision, providing fast distance and gradient queries for manipulation from point-cloud observations.

    \item Using the neural CDF, we formulate a distributionally robust control barrier function (CBF) for safe manipulator control that accommodates multiple sources of uncertainty, including neural CDF modeling errors, point-cloud observation noise, and dynamic obstacle motion.

    \item We develop a sampling-based planner that builds a graph of safe configuration-space bubbles using neural CDF queries, reducing explicit collision checking and improving planning efficiency.

    \item We validate our integrated planning-and-control approach in both simulation and real-robot experiments with a 6-DoF UFactory xArm manipulator operating in dynamic and cluttered environments. We provide an open-source implementation of our method at \url{https://github.com/ExistentialRobotics/NCSB}.
\end{itemize}

\begin{figure*}
    \centering
    \includegraphics[width=0.9\linewidth]{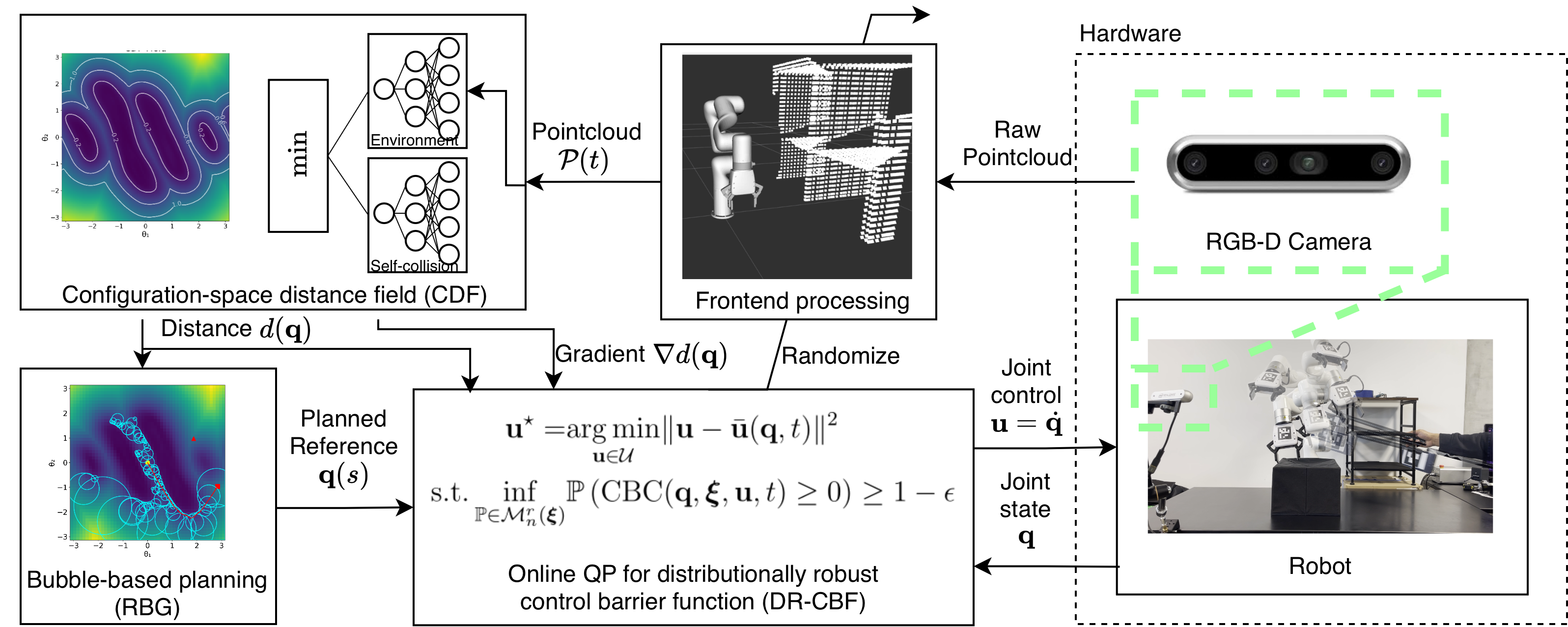}
    \caption{Overview: Two multi-layer perceptron networks, corresponding to the environment and robot self collisions, are trained to approximate the configuration-space distance function (CDF) of a robot manipulator. A sampling-based planner uses CDF values to form collision-free spheres (bubbles) organized in a Rapidly-exploring Bubble Graph (RBG). A controller imposes safety constraints via a distributionally-robust control barrier function obtained from the CDF and solves quadratic programs online to track the planned trajectory safely. The controller considers all distributions within a Wasserstein-1 ball of sampled environment and robot states obtained by randomizing the front-end CDF estimation model.}
    \label{fig:overview}
\end{figure*}

\section{Related Work}
This section reviews prior work on \textit{motion planning} and \textit{safe control} for robotic manipulators.

\noindent
\textbf{Manipulator planning.}  
Sampling-based motion planning algorithms, such as Rapidly-exploring Random Trees (RRT)~\cite{lavalle2001rapidly}, Probabilistic Roadmaps (PRM)~\cite{kavraki1996probabilistic}, and their variants~\cite{rrt_connect_2000, Hauser_2015_planning}, are widely used for generating global trajectories in high-dimensional configuration spaces. These methods have been successful in cluttered or unknown environments and are often paired with low-level feedback controllers, such as PID~\cite{johnson2005pid}, to track the computed paths. However, these planners suffer from high collision-checking overhead and lack real-time adaptability, making them less effective in dynamic or uncertain environments where frequent replanning is needed.

A long-standing challenge in motion planning is how to represent obstacles in the configuration space (\textit{C-space}), which is crucial for generating safe and efficient trajectories. One class of approaches focuses on explicitly approximating the collision-free subset of C-space. Decomposition-based methods such as IRIS~\cite{Deits2015}, and more recent works~\cite{dai2024certified, werner2024faster}, construct a union of convex polyhedra to approximate the free space by solving a series of optimization problems, typically assuming that workspace obstacles are simple geometric primitives (e.g., spheres and cylinders). \cite{verghese_2022_ral} proposed D-Fastron, which uses k-means clustering in forward kinematics space to decompose the configuration space and trains local kernel perceptron models for fast proxy collision checking. Alternative approaches, such as~\cite{adnan_c_space_bubble_icra16,danielson2023invariant}, utilize workspace distances to construct safe regions; however, these methods are restricted to simple manipulator geometries, such as infinitely thin lines. Additionally, safe spherical regions have been explored for efficient collision checking \cite{bialkowski2016efficient,shkolnik2011sample}, biased sampling in free workspace representations \cite{brock2001decomposition}, and navigation function design \cite{yang2004sng}. Another relevant line of work composes invariant funnels of low-level controllers \cite{kdf_kavraki,tedrake2010lqr} to plan a safe sequence of controller switches to a desired goal.
Our approach differs by using spherical regions to construct a rapidly exploring random graph in configuration space, extending the idea introduced by \cite{safe_bubble_cover} to high-dimensional robot configurations and more complex environments. Compared to algorithms that rely on safe regions with explicit collision checking or iterative inflation, our bubbles are constructed using fast batched neural CDF distance queries, reducing expensive collision-check calls. Compared to trajectory optimization over general convex sets that often requires mixed-integer global planning \cite{dai2024certified, werner2024faster}, our approach provides free-space approximation via a union of spheres constraints that support sampling-based planning with fewer collision checks or convex optimization of feasible global trajectories. The planned trajectory is executed by an optimization-based safe tracking controller that enforces the safety constraint online. In addition to simplicity, this decoupling allows accounting for uncertainties (e.g., in obstacle sensing and neural models) online during execution, avoiding the conservatism required to account for all uncertainties offline during planning.

More recently, neural representations have emerged to directly model the geometry of the manipulator and its environment. Signed distance fields (SDFs) are used to encode collision information in the workspace \cite{koptev_neural_jsdf_2022, driess2022learning, zhu2024efficient, li2024representing}, and have been integrated into reactive planning frameworks \cite{vasilopoulos2023ramp} and receding-horizon planners~\cite{michaux2024safe_rss}. While SDFs offer differentiability, they operate in the workspace and often require projection to C-space for planning and control. To overcome this limitation, \cite{li2024cdf_rss} introduced the concept of CDF, which estimates the shortest joint-space distance from a configuration to workspace obstacles using a learned model, which we build on. A recent concurrent paper \cite{probabilistic_bubble_roadmap} also uses a neural CDF to obtain safe regions in configuration space from a workspace represented as a collection of spheres. We consider a more practical setting where the workspace is partially observed with an RGB-D sensor, and augment the planning module with a distributionally-robust controller that avoids collisions using online sensor measurements. These aspects enable the demonstration of our method on a real robot.

\noindent
\textbf{Manipulator safe control.}  
Safe and efficient execution of motion plans requires controllers capable of handling high-dimensional dynamics, enforcing safety constraints, and adapting to changes in the environment. Model predictive control~\cite{richalet1978model} has been widely applied to robotic manipulators due to its ability to incorporate constraints and optimize control actions over a receding horizon~\cite{kleff2021high, dai2020robust}. Sampling-based variants such as model predictive path integral control~\cite{williams2016aggressive} have demonstrated effectiveness in handling non-convex costs and complex dynamics, including for robotic manipulators~\cite{vasilopoulos2023ramp, zhang2024_ral}. In parallel, control barrier functions (CBFs) have emerged as a popular framework for enforcing safety constraints across a range of robotic systems~\cite{long2024sensorbased_dro, keyumarsi_LiDAR_CBF, yu2024_cbf_rrt, zhang2025gcbf+}. CBFs enable the construction of safety filters that guarantee forward invariance of safe sets by solving a real-time quadratic program (QP), making them computationally tractable and easy to integrate with existing nominal control policies.

CBF-based approaches have increasingly been applied to robotic manipulation tasks. \cite{singletary_2023_ral_sdf_cbf_qp} proposed a CBF-based trajectory filtering strategy that dynamically adjusts pre-planned trajectories to account for updated collision constraints. \cite{yu2024_cbf_rrt} developed a CBF-induced neural controller (CBF-INC) that integrates with RRT for motion planning, utilizing state-based or point-cloud inputs to enhance success rates and reduce exploration nodes.\cite{djeha_2023_tro} introduced integral feedback into high-gain task-space QPs to ensure provable closed-loop stability under unmodeled dynamics. \cite{chi2024safe} leveraged neural CDFs as control barrier functions (CBFs) for safe control, demonstrating that CDF-based methods outperform SDF-based approaches due to their more reliable gradient estimates. \cite{brunke2025semantically} proposed a semantic safety filter that integrates scene semantics with CBFs to enforce “common-sense” safety. \cite{morton2025safe} introduced an operational-space CBF formulation that enforces hundreds of simultaneous safety constraints (e.g., collision, singularity, workspace limits) at kilohertz control rates while preserving task consistency via operational-space hierarchy.





\section{Problem Formulation}
\textbf{Notation:} We use $\mathbb{R}$, $\mathbb{R}_{\ge0}$, and $\mathbb{N}$ for the real, nonnegative real, and natural numbers, and $[N]=\{1,\dots,N\}$. For a set $\mathcal{C}\subset\mathbb{R}^n$, $\mathrm{Int}(\mathcal{C})$ and $\partial\mathcal{C}$ denote its interior and boundary. The gradient of a differentiable function $V$ is $\nabla V$. A function $\alpha$ is of class $\mathcal{K}_\infty$ if it is continuous, strictly increasing, and satisfies $\alpha(0)=0$ and $\alpha(r) \to \infty$ as $r\to\infty$. The sets $\text{SO}(p) = \{\bfR \in \bbR^{p \times p} \mid \bfR^\top \bfR = \bfI_p, \det(\bfR) = 1\}$ and $\mathrm{SE}(p)$ with elements $\begin{pmatrix} \mathbf{R} & \mathbf{t} \\ \mathbf{0}^\top & 1 \end{pmatrix} \in \bbR^{(p+1) \times (p+1)}$ where $\mathbf{R} \in \text{SO}(p), \; \mathbf{t} \in \mathbb{R}^p$ denote the special orthogonal and special Euclidean groups, respectively.

\textbf{Problem Statement:} We consider an $m$-DoF revolute manipulator with joint configuration $\mathbf{q}=[q_1,\dots,q_m]^\top\in\mathcal{Q}$, where $\mathcal{Q}:=\{\mathbf{q}\in\mathbb{R}^m\mid \mathbf{q}_{\min}\le\mathbf{q}\le\mathbf{q}_{\max}\}$ denotes the joint-limit–bounded configuration space. The end-effector pose is given by the forward-kinematics map $\mathbf{T}_{\mathrm{ee}}(\mathbf{q})\in\mathrm{SE}(3)$, and the occupied workspace volume of the robot is represented by a set-valued shape function $\mathcal{S}(\mathbf{q})\subset\mathbb{R}^3$ with boundary $\partial\mathcal{S}(\mathbf{q})$. The robot kinematics are:
\begin{equation}\label{eq:dynamics}
\dot{\mathbf{q}}(t)=\mathbf{u}(t),\qquad \mathbf{q}(0)=\mathbf{q}_0,
\end{equation}
where $\mathbf{u}(t)\in\mathcal{U}\subset\mathbb{R}^m$ is the joint-velocity input and $\mathbf{q}_0$ is the initial configuration. The robot operates in a dynamic environment with obstacles $\mathcal{O}(t)\subset\mathbb{R}^3$ and free space $\mathcal{F}(t)=\mathbb{R}^3\setminus\mathcal{O}(t)$. A sensor (e.g., depth camera or LiDAR) provides point-cloud measurements $\mathcal{P}(t)\subset\mathbb{R}^3$ approximating $\partial\mathcal{O}(t)$, and the sensor-to-base extrinsic calibration is assumed known.


\begin{problem}
\label{problem:3d_navigation}
Consider a robot manipulator with initial configuration $\bfq_0$ and dynamics in \eqref{eq:dynamics}. Given point-cloud observations \(\calP(t)\) of the obstacles $\calO(t)$ in the environment, design a control policy that drives the robot's end-effector $\bfT_{\text{ee}}(\bfq(t))$ to a desired goal pose $\mathbf{T}_{\text{G}} \in \text{SE}(3)$, while ensuring the robot body remains in free space, i.e., $\mathcal{S}(\mathbf{q}(t)) \subset \mathcal{F}(t)$, $\forall t \geq 0$.
\end{problem}
 
\section{Safe Barrier from Configuration-Space Distance Function}
\label{sec: safe_barrier_cdf}
Efficient motion planning and control for robotic manipulators require scalable and precise safety representations in configuration space. This section reviews signed distance functions (SDFs), introduces environment and self-collision CDFs and their key properties, and concludes with the definition of a neural CDF barrier for planning and control.

\subsection{Environment Configuration Space Distance Function}
\label{sec: environ_cdf}

Motivated by successful applications in safe mobile robot navigation \cite{Han_2019_fiesta, Long_learningcbf_ral21, keyumarsi_LiDAR_CBF},
SDFs have recently been used for motion planning and control of robot manipulators \cite{koptev_neural_jsdf_2022, li2024representing, yu2024_cbf_rrt,long2024neural_continuum_sdf}. An SDF $f_s$ measures the distance from a workspace point $\mathbf{p} \in \bbR^3$ to the robot surface $\partial \calS(\mathbf{q})$:
\begin{equation}
f_s(\mathbf{p}, \mathbf{q}) =    
\begin{cases}
        -\min\limits_{\mathbf{p}^* \in \partial \calS(\bfq)} \| \mathbf{p} - \mathbf{p}^* \|, & \text{if } \mathbf{p} \in \calS(\bfq), \\
        \phantom{-}\min\limits_{\mathbf{p}^* \in \partial \calS(\bfq)} \| \mathbf{p} - \mathbf{p}^* \|, & \text{if } \mathbf{p} \notin \calS(\bfq).
    \end{cases}
\label{eq: sdf_definition}
\end{equation}
When represented using a differentiable model (e.g., neural network \cite{deepsdf,koptev_neural_jsdf_2022}), this function enables efficient motion planning and collision avoidance by facilitating rapid computation of distances and gradients. While SDFs are widely applied to mobile robots to enforce workspace safety constraints, manipulators with high-dimensional joint spaces benefit from a configuration-centric perspective. \cite{li2024cdf_rss} introduced a configuration distance function (CDF), which encodes the minimum joint-space distance (in radians) needed for a robot at configuration $\mathbf{q}$ to make contact with a point $\mathbf{p}$. Formally, given a robot SDF $f_s$, the robot CDF can be computed as
\begin{equation}
f_c(\mathbf{p}, \mathbf{q}) = \min_{\mathbf{q}'} \| \mathbf{q} - \mathbf{q}' \|, \; \; \text{subject to} \; \; f_s(\mathbf{p}, \mathbf{q}') = 0.
\label{eq: cdf_definition}
\end{equation}
Similar to an SDF, this CDF representation satisfies an Eikonal equation with respect to $\mathbf{q}$:
\begin{equation}
\| \nabla_{\mathbf{q}} f_c(\mathbf{p}, \mathbf{q}) \| = 1,
\label{eq:eikonal_cdf}
\end{equation}
whenever it is differentiable in $\mathbf{q}$.

In practice, computing \eqref{eq: cdf_definition} requires identifying a set of zero-level configurations \(\mathbf{q}'\) satisfying \(f_s(\mathbf{p}, \mathbf{q}') = 0\). These configurations are typically obtained using numerical optimization methods such as the Broyden-Fletcher-Goldfarb–Shanno (BFGS) algorithm \cite{head1985broyden}. However, due to the high dimensionality of the configuration space and the sparsity of \(\mathbf{q}'\) samples, this direct formulation can lead to an overly smooth or inaccurate approximation of the true CDF. Therefore,  \cite{li2024cdf_rss} leverages the fact that the contact at a point \(\mathbf{p}\) is primarily determined by a subset of joints preceding the contact link. Let \(k\) denote the robot link that comes into contact with \(\mathbf{p}\), and let \(\mathbf{q}_{:k}\) represent the joint angles influencing the motion of link \(k\). The CDF is then refined by restricting the distance computation to these relevant joints:
\begin{equation}
f_c(\mathbf{p}, \mathbf{q}) = \min_{k=1,\dots,m} \min_{\mathbf{q}'} \| \mathbf{q}_{:k} - \mathbf{q}'_{:k} \|, \quad \text{s.t.} \quad f_s(\mathbf{p}, \mathbf{q}') = 0,
\label{eq:cdf_refined}
\end{equation}
where \(m\) is the total number of robot joints.

\subsection{Self-Collision Configuration Space Distance Function}
While environment CDFs measure proximity to external obstacles, self-collisions between different parts of the robot pose a distinct challenge, especially for high-DoF manipulators with complex self-collisions. To address this, we define a \textit{Self-Collision Configuration Distance Function} (SCDF), which quantifies the minimum joint-space distance from a configuration \(\mathbf{q}\) to the set of self-colliding configurations.

Let \(\mathcal{C}_{\mathrm{sc}} \subset \calQ\) denote the closed set of joint configurations that result in self-collision. Then, for a given configuration \(\mathbf{q} \in \calQ\), we define the SCDF as:
\begin{equation}
f_{\mathrm{sc}}(\mathbf{q}) := \min_{\mathbf{q}' \in \mathcal{C}_{\mathrm{sc}}} \| \mathbf{q} - \mathbf{q}' \|.
\label{eq:scdf_full}
\end{equation}

Similar to Section~\ref{sec: environ_cdf}, the SCDF in~\eqref{eq:scdf_full} 
represents the Euclidean distance from \(\mathbf{q}\) to the closed set of self-colliding configurations 
\(\mathcal{C}_{\mathrm{sc}} \subset \bbR^m\). Note that \(\mathcal{C}_{\mathrm{sc}}\) may consist of multiple disconnected components, 
as different joint subsets can lead to distinct types of self-collision. Nevertheless, the SCDF inherits the Eikonal property and satisfies
\begin{equation}
\label{eq: scdf_eikonal}
\|\nabla_{\mathbf{q}} f_{\mathrm{sc}}(\mathbf{q})\| = 1 . 
\end{equation}

In practice, \(\mathcal{C}_{\mathrm{sc}}\) is often only sparsely sampled, and self-collisions typically involve only a subset of joints. As such, computing the full-joint distance in \eqref{eq:scdf_full} can lead to overly conservative and inaccurate approximations. To improve accuracy and better exploit the robot's kinematic structure, we refine the SCDF by computing distance only over the subset of joints responsible for the self-collision.

Specifically, if a self-collision at \(\mathbf{q}'\) occurs between Link \(i\) and Link \(j\), we define the joint subvector \(\mathbf{q}_{a:b}\), where \(a = \min(i, j)\) and \(b = \max(i, j)\), to include only joints that affect the relative motion of these links. The refined SCDF becomes:
\begin{equation}
f_{\mathrm{sc}}(\mathbf{q}) := \min_{\mathbf{q}' \in \mathcal{C}_{\mathrm{sc}}} \| \mathbf{q}_{a:b} - \mathbf{q}'_{a:b} \|.
\label{eq:scdf_def}
\end{equation}
For example, if the collision occurs between Link 3 and Link 5, then joints 3 through 5 may primarily determine their relative positions, so the distance is computed using \(\mathbf{q}_{3:5}\).

\subsection{Neural CDF Barrier}
\label{subsec: neural_cdf_barrier}
Building on the environment and self-collision CDF, we define a CDF barrier to ensure the robot remains in a collision-free region of the configuration space. Formally, a CDF barrier associated with environment CDF $f_c$, self-collision CDF $f_{sc}$, and obstacle set $\calO(t)$ is defined as:
\begin{equation}
    h(\mathbf{q}, t) := \min\{\inf_{\mathbf{p} \in \partial\mathcal{O}(t)} f_c(\mathbf{p}, \mathbf{q}), f_{sc}(\bfq)\} 
    \label{eq: cdf_barrier}
\end{equation}
where $\partial \calO(t)$ is the obstacle set boundary. Then, the time-varying \emph{safe set} in configuration space induced by $h(\bfq,t)$ is
\begin{equation}
    \mathcal{C}_{\text{safe}}(t) = \{ \mathbf{q} \in \mathcal{Q} \mid h(\mathbf{q}, t) \geq 0 \}.
    \label{eq: cdf_safe_set}
\end{equation}
%


In real-world settings, $\partial \mathcal{O}(t)$ must be inferred from point-cloud measurements $\calP(t)$ from a depth camera or LiDAR. Because the points in $\calP(t)$ are noisy samples from the boundary of $\calO(t)$, we approximate the CDF barrier as
\begin{equation} \label{eq: cdf_barrier_sensor}
    h(\bfq, t) 
    \;\approx\; 
    \min \{\min_{\bfp \,\in\, \mathcal{P}(t)} f_c(\mathbf{p}, \mathbf{q}), f_{sc}(\bfq) \}.
\end{equation}
Moreover, computing $f_c(\mathbf{p}, \mathbf{q})$ and $f_{sc}(\bfq)$ exactly is difficult for high-DoF manipulators because it involves (infinitely) many potential contact configurations $\mathbf{q}'$ in \eqref{eq: cdf_definition} and self-collision configurations in \eqref{eq:scdf_def}. Similar to \cite{li2024cdf_rss}, we model $f_c$ and $f_{sc}$ by multi-layer perceptrons $\hat{f}_c(\mathbf{p}, \mathbf{q}; \boldsymbol{\theta_1})$ and $\hat{f}_{sc}(\mathbf{p}, \mathbf{q}; \boldsymbol{\theta_2})$ with learnable parameters $\boldsymbol{\theta_1}$ and $\boldsymbol{\theta_2}$, respectively. This representation reduces storage requirements compared to volumetric or tabular encodings and supports parallel distance or gradient queries for a set of points $\bfp$ or configurations $\bfq$. Combining the point-cloud measurements with the neural CDF model yields a practical CDF barrier:
\begin{equation} \label{eq: neural_cdf_barrier}
    \hat{h}(\bfq, t; \bftheta)
    \; = \;
    \min \{ \min_{\bfp \,\in\, \mathcal{P}(t)}
    \hat{f}_c(\bfp, \bfq;\, \boldsymbol{\theta_1}), \hat{f}_{sc}(\bfq; \boldsymbol{\theta_2})\}.
\end{equation}
However, learning-based representations inevitably introduce modeling errors. The noise in $\mathcal{P}(t)$, its lack of complete coverage of the obstacle boundary, and the approximation error in $\hat{f}_c$ and $\hat{f}_{sc}$ constitute multiple sources of uncertainty. 


In what follows, we leverage the fast parallel query capabilities of the neural CDF barrier to enable efficient sampling-based planning in configuration space (Sec.~\ref{sec: bubble-based planning}), and propose a distributionally robust control strategy that tracks the planned trajectory while ensuring real-time safety despite the aforementioned uncertainties and motion in obstacles (Sec.~\ref{sec: dro_cbf_safety}).

\section{Configuration Space Bubble-Based Planning}
\label{sec: bubble-based planning}

Motion planning for high-DoF manipulators typically involves numerous collision checks \cite{Hauser_2015_planning}, making global motion planning computationally expensive. To address this, we propose the \textbf{bubble-CDF planner}, a sampling-based approach that efficiently explores the configuration space while ensuring safety via neural CDF barriers. 
Unlike conventional RRT-based planners that rely on dense edge validation, our approach constructs \textit{configuration space bubbles} as local safe regions, enabling rapid exploration and collision-free path generation. 

In Problem~\ref{problem:3d_navigation}, the robot must move from an initial configuration $\mathbf{q}_0$ to any valid joint configuration $\mathbf{q}_\text{G}$ that achieves a desired end-effector pose $\bfT_{\text{ee}}(\bfq_\text{G}) = \mathbf{T}_\text{G} \in \mathrm{SE}(3)$. 
Inverse kinematics (IK) provides $K$ goal configurations $\{\mathbf{q}_\text{G}^{i}\}_{i \in [K]}$, each of which achieves the desired end-effector pose. 

Our approach builds on the \textit{Rapidly-exploring Bubble Graph (RBG)} algorithm~\cite{safe_bubble_cover}, which constructs a configuration-space roadmap where each node represents a locally safe region (bubble) instead of a single configuration. The learned CDF barrier \(\hat{h}(\mathbf{q}, t)\) is used to define the bubble radius dynamically, ensuring rapid, collision-free exploration without incremental edge validation. This substantially reduces the collision-checking overhead in high-dimensional spaces while preserving the efficiency of sampling-based planners. 

\subsection{Rapidly-exploring Bubble Graph (RBG)}
Suppose $\bfq \in \calQ$ is a safe configuration. We aim to identify a radius $r(\bfq) \geq 0$ such that any configuration $\bfq'$ satisfying $\|\bfq - \bfq'\| \leq r(\bfq)$ is safe. 
We call this spherical region a \emph{configuration-space bubble} and denote it as $\mathcal{B}(\bfq, r(\bfq))$. Given the neural CDF barrier $\hat{h}(\bfq, t)$ for fixed time $t$, we construct the bubble radius at configuration $\bfq$ as: 
\begin{equation}
\label{eq: cdf_bubble}
    r(\bfq) = \hat{h}(\bfq, t) - \eta, 
\end{equation}
where $\eta > 0$ is a safety margin. All configurations $\bfq'$ within this radius satisfy the neural CDF barrier with margin $h(\bfq', t) \geq \eta$~\cite{safe_bubble_cover}.



To explore a high-dimensional configuration space efficiently using these local safety certificates, RBG adapts the RRT algorithm \cite{lavalle2001rapidly} so that each vertex in the graph corresponds to a bubble rather than a single configuration. In RBG, the step size, corresponding to the distance the planned tree advances toward a sampled point, is dynamically set by the radius of each bubble. This obviates manual tuning of a fixed step size and reduces collision-checking overhead.




Algorithm~\ref{alg:bubble_graph}
details how RBG constructs a bubble-based graph. 
RBG iteratively samples random configurations (biased toward goal configurations), identifies the nearest bubble in configuration space, and if the sample lies outside that bubble, creates a new bubble by querying the neural CDF barrier to determine its radius. The newly created bubble is then connected to any existing bubbles $\calB(\bfq_i, r(\bfq_i))$ with overlapping safe regions, i.e., if 
\begin{equation}
\|\mathbf{q}_{\mathrm{new}} - \mathbf{q}_i\| 
\;\le\;
r(\mathbf{q}_{\mathrm{new}}) + r(\mathbf{q}_i).
\end{equation}
This process continues until the maximum number of bubbles $N_{\mathrm{max}}$ or a goal configuration is reached.

\begin{algorithm}[t]
\caption{Rapidly Exploring Bubble Graph}
\label{alg:bubble_graph}
\small
\begin{algorithmic}[1]
\Require Start configuration $\mathbf{q}_{0}$, goal configurations $\{\mathbf{q}_{\mathrm{G}}^i\}_{i=1}^K$, Neural CDF barrier $\hat{h}(\mathbf{q}, t)$, safety margin $\eta$, Max. no. of bubbles $N_{\mathrm{max}}$, Min. radius $r_{\mathrm{min}}$.
\Ensure Graph $\mathcal{G}$ containing bubbles from start to goals
\Function{BuildBubbleGraph}{$\mathbf{q}_0,\{\mathbf{q}_{\mathrm{G}}^i\}$}
    \State $\mathcal{G}.V \gets \{\mathcal{B}(\mathbf{q}_{0}, \hat{h}(\mathbf{q}_{0}, t) - \eta)\}, \quad \mathcal{G}.E \gets \emptyset$
    
    \While{$|\mathcal{G}.V| < N_{\mathrm{max}}$}
        \State $\mathbf{q}_{\mathrm{rand}} \gets \Call{SampleRandom}{}()$ \Comment{With goal bias}
        \State $\mathcal{B}_{\mathrm{near}} \gets \Call{NearestBubble}{\mathcal{G}, \mathbf{q}_{\mathrm{rand}}}$

        \If{$\mathbf{q}_{\mathrm{rand}}$ \textbf{outside} $\mathcal{B}_{\mathrm{near}}$}
            \State $\mathbf{q}_{\mathrm{new}} \gets \Call{ExtendToward}{\mathcal{B}_{\mathrm{near}}, \mathbf{q}_{\mathrm{rand}}}$
            \State $r_{\mathrm{new}} \gets \hat{h}(\mathbf{q}_{\mathrm{new}}, t) - \eta$
            \If{$r_{\mathrm{new}} > r_{\mathrm{min}}$}
                \State $\mathcal{B}_{\mathrm{new}} \gets \mathcal{B}(\mathbf{q}_{\mathrm{new}},r_{\mathrm{new}})$
                \State \Call{UpdateConnections}{$\mathcal{G},\mathcal{B}_{\mathrm{new}}$}
            \EndIf
        \EndIf
    \EndWhile
    \State \Return $\mathcal{G}$
\EndFunction

\Function{UpdateConnections}{$\mathcal{G},\mathcal{B}_{\mathrm{new}}$}
    \For{$\mathcal{B}_i \in \mathcal{G}.V$}
        \If{$\|\mathcal{B}_{\mathrm{new}}.\mathbf{q}-\mathcal{B}_i.\mathbf{q}\|\le r(\mathbf{q}_{\mathrm{new}})+r(\mathbf{q}_i)$}
            \State $\mathcal{G}.E \gets \mathcal{G}.E \,\cup\, (\mathcal{B}_{\mathrm{new}},\mathcal{B}_i)$
        \EndIf
    \EndFor
    \State $\mathcal{G}.V \gets \mathcal{G}.V \,\cup\, \{\mathcal{B}_{\mathrm{new}}\}$
\EndFunction
\end{algorithmic}
\end{algorithm}

\subsection{Path Selection and Trajectory Optimization}
\label{subsec:path_optimization}
Once the bubble graph $\mathcal{G}$ is constructed, we must extract and refine a collision-free path connecting the start configuration $\mathbf{q}_{0}$ to one of the $K$ goal configurations $\{\mathbf{q}_{\mathrm{G}}^i\}$. Since $\mathcal{G}$ may contain multiple feasible routes, we first select an optimal path according to a desired criterion (e.g., minimum total edge distance). We then solve a local trajectory optimization problem, fitting a continuous curve through the sequence of bubbles to ensure a smooth, dynamically feasible trajectory that remains collision-free. 

\textbf{Path Selection.}
We assign a cost to each edge in $\mathcal{G}$, for instance, the single-sided Hausdorff distance between two overlapping bubbles~\cite{safe_bubble_cover}. A shortest-path search (e.g., Dijkstra \cite{DIJKSTRA1959} or A* \cite{A_star_planning}) is run from the start bubble to every goal bubble. The path with the lowest overall cost is then passed to the optimization step, as summarized in Algorithm~\ref{alg:path_select_opt}.

\textbf{Trajectory Optimization.}
To obtain a smooth trajectory, we optimize a sequence of Bézier curves connecting the start configuration $\bfq_0$ to the selected goal configuration $\bfq_\text{G}^\star$ through the path of bubbles $\{\mathcal{B}_1, \ldots, \mathcal{B}_n\}$. Each Bézier curve segment of degree $d$ is defined by its control points:
\begin{equation}
\gamma_j(t) = \sum_{l=0}^d \binom{d}{l} t^l(1-t)^{d-l} \mathbf{c}_{j,l},
\end{equation}
where $\mathbf{c}_{j,l} \in \mathbb{R}^m$ is the $l$-th control point of the $j$-th curve segment. The $k$-th derivative of a Bézier curve is also a Bézier curve, and its squared norm integral can be expressed as: 
\begin{equation}
\int_0^1 \|\gamma_j^{(k)}(t)\|^2 dt = \mathbf{c}_j^\top \mathbf{Q}_k \mathbf{c}_j,
\end{equation}
where $\mathbf{c}_j$ is the vectorized form of all control points in segment $j$, and $\mathbf{Q}_k$ is a positive semidefinite matrix. This leads to the following convex quadratically constrained quadratic program:
\begin{equation}
\begin{aligned}
\label{eq: bezier_opti}
\min_{\{\mathbf{c}_{j,i}\}} \quad & \sum_{j=1}^n \sum_{k=1}^K w_k \int_0^1 \|\gamma_j^{(k)}(t)\|^2 dt \\
\text{subject to} \quad & \|\mathbf{c}_{j,l} - \mathbf{q}_j\|^2 \leq r(\mathbf{q}_j)^2, 
    \quad \forall j,l \\
& \gamma_j(1) = \gamma_{j+1}(0), 
    \quad j = 1,\ldots,n-1 \\
& \gamma_j^{(k)}(1) = \gamma_{j+1}^{(k)}(0), 
    \quad j = 1,\ldots,n-1, \; k = 1,2 \\
& \gamma_1(0) = \bfq_0, \quad \gamma_n(1) = \bfq_\text{G}^*, \\
& \gamma_1^{(i)}(0) = \mathbf{0}, \quad \gamma_n^{(i)}(1) = \mathbf{0}, \quad i = 1,\ldots,d 
\end{aligned}
\end{equation}
where $\mathbf{c}_{j,i}$ are the control points of the $j$-th Bézier curve segment $\gamma_j$ of degree $d$, and $\gamma_j^{(k)}$ denotes its $k$-th derivative. The objective minimizes a weighted sum of integrated squared derivatives, promoting smoothness. The first constraint ensures safety by keeping control points within their respective bubbles, which, by the convex hull property of Bézier curves, guarantees the entire trajectory remains collision-free. The remaining constraints enforce continuity of position and derivatives between segments, as well as boundary conditions including zero velocity at endpoints. Unlike traditional sampling-based planners that may not have safety guarantees or require additional collision checks for a post-optimized trajectory, our formulation provides a convex optimization problem that simultaneously guarantees smoothness and safety without requiring additional collision checking.





\begin{algorithm}[t]
\caption{Path Selection and Trajectory Optimization}
\label{alg:path_select_opt}
\small
\begin{algorithmic}[1]
\Require
\State Graph $\mathcal{G}$ with start bubble $\mathcal{B}_{\mathrm{start}}$, goal bubbles $\{\mathcal{B}_{\mathrm{goal}}^i\}_i$ 
\State Edge cost function $\textsc{EdgeCost}(\mathcal{B}_i,\mathcal{B}_j)$ 
\State Weights $\{w_k\}$ for trajectory cost 
\Ensure Smooth collision-free Bézier trajectory from $\mathbf{q}_{0}$ to some $\mathbf{q}_{\mathrm{G}}^i$
\Function{PathSelectAndOptimize}{$\mathcal{G}, \mathcal{B}_{\mathrm{start}}, \{\mathcal{B}_{\mathrm{goal}}^i\}$}
    \State $\textit{bestCost} \gets \infty,\quad \textit{bestPath} \gets \textit{None},\quad i_{\mathrm{best}} \gets -1$
    \For{$i=1 \to K$}
        \State $\textit{path}_i \gets \Call{ShortestPath}{\mathcal{G}, \mathcal{B}_{\mathrm{start}}, \mathcal{B}_{\mathrm{goal}}^i, \textsc{EdgeCost}}$
        \If{$\textit{path}_i$ exists}
            \State $\textit{cost}_i \gets \Call{PathCost}{\textit{path}_i, \textsc{EdgeCost}}$
            \If{$\textit{cost}_i < \textit{bestCost}$}
                \State $\textit{bestCost} \gets \textit{cost}_i$, $\textit{bestPath} \gets \textit{path}_i$,  $i_{\mathrm{best}} \gets i$
            \EndIf
        \EndIf
    \EndFor

    \If{$\textit{bestPath} = \textit{None}$}
        \State \Return \textit{failure} \Comment{No feasible path found}
    \EndIf
    \State $\textbf{curves} \gets \Call{BezierOpti}{\textit{bestPath}, \{w_k\}}$ \Comment{Eq. \eqref{eq: bezier_opti}}
    \State \Return $\bigl(\textbf{curves},\, i_{\mathrm{best}}\bigr)$
\EndFunction
\end{algorithmic}
\end{algorithm}

Throughout this section, we have introduced the \textbf{bubble-CDF planner}, which leverages neural CDF barriers to efficiently explore the configuration space while ensuring safety. By constructing \textit{configuration-space bubbles} around safe configurations and dynamically adjusting step sizes based on bubble radii, our approach significantly reduces collision-checking overhead. Additionally, the trajectory optimization formulation in \eqref{eq: bezier_opti} guarantees smooth, dynamically feasible paths without requiring post-processing for collision checks.

Thus far, we have described the bubble--CDF planner in a nominal setting, where the neural CDF barrier $\hat{h}(\mathbf{q}, t)$ is treated as time‐invariant and accurate. In real deployments, however, dynamic obstacles, partial/noisy point-cloud observations, CDF approximation errors, and sensor noise may cause the bubble radius to be locally overestimated, leading to a nominally collision-free plan under $\hat{h}$ that is unsafe in the true environment. To address these uncertainties, Sec.~\ref{sec: dro_cbf_safety} introduces a distributionally robust control barrier function that accounts for potential CDF errors and planning over-approximations, and enforces safety at execution time by adjusting the nominal trajectory when needed.

\section{Distributionally Robust Manipulation Control}
\label{sec: dro_cbf_safety}

In this section, we present a control synthesis approach that guarantees safe real-time configuration trajectory tracking using neural CDF barrier estimates subject to sensing and modeling errors, and even dynamic obstacles. We first describe our approach to track the reference trajectory (Sec.~\ref{subsec:clf_tracking}), and incorporate a distributionally robust CBF (DR-CBF)~\cite{long2024sensorbased_dro} to ensure safe control synthesis despite dynamic obstacles and neural CDF barrier uncertainty (Sec. \ref{subsec: dr_cbf}).


\subsection{Governor-based Trajectory Tracking}
\label{subsec:clf_tracking}

We first present our approach for tracking the planned piecewise Bézier configuration trajectory $\gamma: [0, 1] \to \calQ$ from Sec.~\ref{subsec:path_optimization}, in the absence of safety constraints. Inspired by \cite{garone2015explicit, li2020fast}, we design a reference governor to determine a time-varying reference configuration \(\gamma(s)\) along the trajectory, where \(s \in [0, 1]\) is a progress variable. In other words, $\gamma(0) = \bfq_0$ is the initial robot configuration, while $\gamma(1) = \bfq_{\text{G}}$ is one of the goal configurations. The governor system adapts the value of the progress variable $s$ based on the manipulator's current configuration to ensure smooth progress along the trajectory. We then stabilize the manipulator configuration $\bfq$ to \(\gamma(s)\) using a proportional-derivative (PD) controller. 

The dynamics of \(s\) are designed to balance the robot's tracking capabilities with smooth progression along the trajectory:
\begin{equation}
\label{eq: governor_dynamics}
    \dot{s} = \frac{k}{1 + \|\mathbf{q} - \gamma(s)\|} (1 - s^\zeta),
\end{equation}
where \(k > 0\) is a gain parameter, \(\zeta \geq 1\) controls the rate of saturation as \(s\) approaches 1, and \(\|\mathbf{q} - \gamma(s)\|\) is the Euclidean distance between the current and reference configuration. This adaptive mechanism accelerates \( s \) when the manipulator is near the reference and decelerates it when farther away, ensuring smooth progression without abrupt jumps.

To track the reference trajectory \(\gamma\), we use a PD controller as the nominal controller to reach the local goal \(\gamma(s(t))\):
%
\begin{equation}
\label{eq: pd_control}
    \bar{\bfu}(\bfq, t) = -K_P (\mathbf{q} - \gamma(s(t))) - K_D \dot{\mathbf{q}},
\end{equation}
where \(K_P\) and \(K_D\) are positive definite matrices controlling the proportional and derivative contributions, respectively. The first term ensures that the robot is drawn towards \(\gamma(s)\), while the second term provides damping to minimize oscillations.

\subsection{Distributionally Robust Neural CDF Barrier}
\label{subsec: dr_cbf}




To ensure safe operation while tracking the planned trajectory, the robot's configuration \(\bfq\) must remain within the safe set \(\mathcal{C}_{\text{safe}}(t)\) as defined in \eqref{eq: cdf_safe_set}. We utilize the CDF barrier as control barrier functions (CBFs) \cite{cbf_theory} by considering its time derivative. This leads to the following time-varying control barrier condition (CBC), 
\begin{align}
\label{eq: cbc_explicit}
\text{CBC}(\bfq, \bfu, t) &\!:=\! \nabla_{\mathbf{q}} h(\mathbf{q}, t)^\top \mathbf{u} \!+\! \frac{\partial h(\mathbf{q}, t)}{\partial t} \!+\! \alpha_h(h(\mathbf{q}, t)) \\
& = \underbrace{[\nabla_{\bfq} h(\bfq, t)^\top, \; \; \alpha_h(h(\bfq, t)), \; \; \frac{\partial h(\bfq,t)}{\partial t}]}_{\bfxi^\top(\mathbf{q}, t)} \begin{bmatrix} \bfu\\1\\1
\end{bmatrix} \notag , 
\end{align}
where $\alpha_h$ is a class $\calK_{\infty}$ function.

With the nominal PD control law $\bar{\bfu}(\bfq, t)$ in \eqref{eq: pd_control}, one can enforce safety by solving CBF quadratic program (QP) \cite{ames2016control} that generates the safe manipulator velocity control: 
\begin{equation}
\begin{aligned}
\label{eq: clf_cbf_qp}
    \mathbf{u}^* = &\argmin_{\mathbf{u} \in \mathcal{U}} \|\mathbf{u} - \bar{\bfu}(\bfq ,t)\|^2 \\
    \text{s.t.} \; & \; \text{CBC}(\bfq, \bfu, t) \geq 0 ,    
\end{aligned}
\end{equation}
where $\calU := \{\bfu_{\min} \leq \bfu \leq \bfu_{\max}\}$ defines the admissible control space, ensuring the commanded manipulator velocities remain within feasible operational limits.
However, the CDF barrier \(h(\mathbf{q}, t)\) and its gradients are not known exactly but rather are approximated through the neural CDF barrier \(\hat{h}(\mathbf{q}, t)\) in \eqref{eq: neural_cdf_barrier}. Due to uncertainties introduced by the neural network model and the point-cloud data, $\hat{h}$ cannot be treated as an exact representation of $h$. Instead, we work with samples of the associated uncertainty vector $\{\boldsymbol{\xi}_i\}_{i \in [N]}$, defined in \eqref{eq: cbc_explicit}, which capture the variations in $\hat{h}$, its gradient, and its time derivative. The practical approach of obtaining such samples is discussed in Sec.~\ref{subsec: dro_sample_select}.

Drawing inspiration from recent advances in distributionally robust optimization (DRO) \cite{Esfahani2018DatadrivenDR, long2024sensorbased_dro}, we reformulate the CBC in a distributionally robust manner: 
\begin{equation}
\label{eq: dro_cbf_constraint}  \inf_{\mathbb{P} \in \mathcal{M}_N^r} \mathbb{P} \left( \text{CBC}(\mathbf{q}, \mathbf{u}, \boldsymbol{\xi}) \geq 0 \right) \geq 1 - \epsilon,
\end{equation}
where \(\mathcal{M}_N^r\) is a Wasserstein ambiguity set constructed from \(\{\boldsymbol{\xi}_i\}_{i=1}^N\) and $\epsilon$ is a user-specified risk tolerance. To formally define \(\mathcal{M}_N^r\), let \(\Xi \subseteq \mathbb{R}^{m+2}\) denote the space of uncertainty vectors \(\boldsymbol{\xi}\), and let \(\mathcal{P}_p(\Xi)\) be the space of all Borel probability measures over \(\Xi\) with finite \(p\)-th moment, by utilizing the $p$-Wasserstein distance $W_p$ \cite{Esfahani2018DatadrivenDR}, the ambiguity set 
\begin{equation}
    \calM_{N}^{r} := \{\mu \in \calP_p(\Xi) \; | \; W_p(\mu,\mathbb{P}_{N} ) \leq r \}
\end{equation}
defines a ball of distributions around the discrete empirical distribution $\mathbb{P}_N :=  \frac{1}{N}\sum_{i=1}^N \delta_{\bfxi_i}$ of the samples. Intuitively, the $p$-Wasserstein distance measures the minimum amount of probability mass transport needed to transform one distribution into another. Thus, the ambiguity set $\calM_N^r$ encodes all distributions that are within a radius $r$ of the empirical distribution. and offers a principled way to model uncertainty in the neural CDF and point-cloud observations through its parameters (radius $r$ and risk level $\epsilon$), without requiring explicit knowledge of the true distribution of $\boldsymbol{\xi}$.

Next, by utilizing the $1$-Wasserstein distance and following a convex reformulation of the DR-CBF in \cite[Proposition 6.5]{long2024sensorbased_dro}, lead to the following optimization problem:
\begin{align}
\label{eq: clf_dr_cbf_qp}
\mathbf{u}^* = &\argmin_{\bfu \in \calU, s \in \bbR, \beta_i \in \bbR} \|\mathbf{u} - \bar{\bfu}(\bfq ,t)\|^2 , \\
    \text{s.t.} \;
& \; r \|\mathbf{u}\|_\infty \leq s\epsilon - \frac{1}{N}\sum_{i=1}^N \beta_i, \notag \\
& \; \beta_i \geq \max\{s \!-\! [\bfu \; \; 1 \; \; 1]^\top \bfxi_i(\bfq, t), 0\},  \;  \forall i \in [N]. \notag
\end{align}
The reformulation maintains the linear dependence of the constraints on \(\bfu\), ensuring that the optimization remains a QP and can be solved efficiently in real time. 

As with any CBF QP, feasibility is not guaranteed in all situations. Prior work on distributionally robust safety filtering \cite{PM-KL-NA-JC:23-csl} and on compositions of CBFs under input constraints \cite{breeden2023robust} analyzes conditions under which infeasibility can arise, such as restrictive control limits, adversarial obstacle motion, or overly conservative uncertainty sets. In practice, we choose the Wasserstein radius and the number of uncertainty samples so that the DR-CBF constraint remains compatible with the admissible control bounds, and we did not observe infeasibility in any of our experiments.

Our formulation assumes access to a velocity-control interface, which is standard for many robotic manipulators where a low-level torque controller tracks commanded joint velocities. Although the full manipulator dynamics are second-order, the closed-loop behavior at this interface can be written as:
\begin{equation}
\dot{\mathbf{q}} = \mathbf{u} + \mathbf{d}(\mathbf{q},\dot{\mathbf{q}}),
\end{equation}
where $\mathbf{d}(\mathbf{q},\dot{\mathbf{q}})$ represents bounded discrepancies arising from inertia, Coriolis and centrifugal effects, gravity, friction, and other unmodeled dynamics. These discrepancies enter the barrier derivative as additional uncertain terms and can be incorporated into the uncertainty vector $\boldsymbol{\xi}(\mathbf{q},t)$ in \eqref{eq: cbc_explicit}.

Our DR-CBF formulation also extends directly to torque-level control of second-order manipulator dynamics. Defining the state as $\mathbf{x} = (\mathbf{q},\dot{\mathbf{q}})$ and the input as the joint torque $\mathbf{u} = \boldsymbol{\tau}$ yields the standard control-affine representation:
\begin{equation}
\dot{\mathbf{x}} = f(\mathbf{x}) + G(\mathbf{x})\,\mathbf{u},
\label{eq:second_order_affine}
\end{equation}
\begin{equation*}
f(\mathbf{x}) \!=\!
\begin{bmatrix}
\dot{\mathbf{q}} \\
-\mathbf{M}(\mathbf{q})^{-1}\!\left( \mathbf{C}(\mathbf{q},\dot{\mathbf{q}})\dot{\mathbf{q}} \!+\! \mathbf{g}(\mathbf{q}) \right)
\end{bmatrix},
\;
G(\mathbf{x}) \!=\!
\begin{bmatrix}
\mathbf{0} \\
\mathbf{M}(\mathbf{q})^{-1}
\end{bmatrix}.
\end{equation*}
Our DR-CBF approach can be applied to this second-order model to enforce safety at the torque level.

\subsection{Uncertainty Sample Selection}
\label{subsec: dro_sample_select}

This section details our approach for obtaining samples \(\{\boldsymbol{\xi}_i\}_{i=1}^N\) of the uncertainty vector $\bfxi$ in the CBC in \eqref{eq: cbc_explicit}. 
The definition of the neural CDF barrier \(\hat{h}(\mathbf{q}, t)\) in \eqref{eq: neural_cdf_barrier} depends on the point-cloud data \(\mathcal{P}(t)\), which is represented as \(\mathcal{P}(t) = \{\bfp_j\}_{j=1}^{M_1}\). Each \(\bfp_j \in \mathbb{R}^3\) corresponds to a workspace point captured at time \(t\). The velocity of each point, denoted as \(\frac{\partial \bfp_j}{\partial t}\), is also considered to account for dynamic obstacles.

To account for uncertainty in the neural CDF model $\hat{f}(\bfp, \bfq)$, we employ Monte Carlo (MC) dropout in the network architecture \cite{2015yarin_dropout}. MC dropout provides an efficient approximate Bayesian inference method by treating dropout as sampling from an implicit distribution over the network weights.
This allows us to perform \(M_2\) stochastic forward passes through the robot CDF neural network, collecting multiple realizations of \(\hat{f}_c(\mathbf{p}, \mathbf{q}; \boldsymbol{\theta})\), along with its gradients. Specifically, for each forward pass of \(f_c^{(i)}\), we evaluate \(f_c^{(i)}\) across all \(M_1\) points in \(\mathcal{P}(t)\). Consequently, we obtain \(M_1 \times M_2\) realizations of:
\begin{equation}
\label{eq: uncertianty_sample_set}
\left\{f_c^{(i)}(\bfp_j, \mathbf{q}), \nabla_{\mathbf{q}} f_c^{i}(\bfp_j, \mathbf{q}), \nabla_{\mathbf{p}} f_c^{(i)}(\mathbf{p}, \mathbf{q})^\top \frac{\partial \bfp_j}{\partial t}\right\},
\end{equation}
where \(i \in [M_2]\) corresponds to the MC dropout realization and \(j \in [M_1]\) corresponds to the point in \(\mathcal{P}(t)\). 

From the \(M_1 \times M_2\) realizations, we select $N \ll M_1 \times M_2$ samples in the set \eqref{eq: uncertianty_sample_set} based on the $N$ minimal values of  
\begin{equation}
    \alpha_h(f_c^{(i)}(\bfp_j, \mathbf{q})) +  \nabla_{\mathbf{p}} f_c^{(i)}(\mathbf{p}, \mathbf{q})^\top  \frac{\partial \bfp_j}{\partial t} ,
\end{equation}
which effectively identifies samples $\{\bfxi_i\}_{i \in [N]}$. 

This section introduced a DR-CBF formulation that ensures safe manipulation under uncertainty. Unlike naive safety margin adjustments, our approach explicitly accounts for uncertainties in the neural CDF model, its gradient estimates, and point-cloud velocity observations. By leveraging a Wasserstein ambiguity set, we optimize for the worst-case uncertainty within a data-driven confidence region, providing stronger safety guarantees than standard CBF formulations. This enables adaptive and robust control in dynamic environments without requiring explicit probabilistic models of uncertainty.




\section{Evaluation}
\label{sec:evaluation}

This section evaluates our planning and control techniques, which utilize neural CDF barriers, in simulations on a 2-DoF planar arm and the 6-DoF xArm, as well as in real-world experiments on the 6-DoF xArm. We evaluate the computational efficiency and solution quality of the bubble-CDF planner and the robustness of the DR-CBF controller in ensuring safety under dynamic obstacles and uncertainty.

\subsection{Implementation Details}

\textbf{Training Dataset and Loss Function:} We discretize the robot's workspace into a $40 \times 40 \times 40$ grid. For each workspace point, we generate contact configurations using BFGS optimization, and then select 200 diverse configurations via farthest-point sampling in configuration space \cite{li2024cdf_rss}. During training, we employ an online batching strategy where each batch randomly samples 10 configurations and 50 workspace points. For each configuration–point pair \((\mathbf{p}_i, \mathbf{q}_i)\), the ground-truth environment CDF value \(f_c^\star(\mathbf{p}_i, \mathbf{q}_i)\) is computed as the minimum joint-space distance to the set of contact configurations associated with \(\mathbf{p}_i\).

The network is trained using the following loss function: 
\begin{align}
\mathcal{L}_{\text{env}} = \frac{1}{N_1 N_2} \sum_{i=1}^{N_1} &\sum_{j=1}^{N_2}\Big( \hat{f}_c(\mathbf{p}_i, \mathbf{q}_j; \boldsymbol{\theta}_1) - f_c^\star(\mathbf{p}_i, \mathbf{q}_j) \Big)^2 \quad \nonumber \\
+ &\lambda \left( \| \nabla_{\mathbf{q}} \hat{f}_c(\mathbf{p}_i, \mathbf{q}_j; \bftheta_1) \| - 1 \right)^2,
\label{eq:env_cdf_loss}
\end{align}
where  \(\lambda=0.05\) is a user-specified weight.

For the self-collision CDF (SCDF), we generate training data by uniformly sampling joint configurations and detecting self-collisions using the PyBullet simulation engine \cite{coumans2021_pybullet}. Each self-colliding configuration \(\mathbf{q}' \in \mathcal{C}_{\mathrm{sc}}\) is stored along with the minimal joint index range \([a:b]\) responsible for the collision. During training, each batch consists of \(N\) randomly sampled configurations \(\{\mathbf{q}_j\}_{j=1}^N\). For each configuration, the ground-truth SCDF value \(f_{\mathrm{sc}}^\star(\mathbf{q}_j)\) is computed according \eqref{eq:scdf_def}.

The SCDF network is trained with the following loss:
\begin{align}
\mathcal{L}_{\text{sc}} = \frac{1}{N_3} \sum_{j=1}^{N_3} \Big( \hat{f}_{\mathrm{sc}}(\mathbf{q}_j; \boldsymbol{\theta}_2) - f_{\mathrm{sc}}^\star(\mathbf{q}_j) \Big)^2 \quad \nonumber \\
+ \lambda \left( \| \nabla_{\mathbf{q}} \hat{f}_{\mathrm{sc}}(\mathbf{q}_j; \boldsymbol{\theta}_2) \| - 1 \right)^2,
\label{eq:scdf_loss}
\end{align}
where \(\lambda = 0.05\) is user-specified weight.

\textbf{Network Architecture and Training:}
The neural CDF and SCDF are both implemented as 5-layer fully connected multilayer perceptrons (MLPs) with GELU activation. Skip connections are included at the second and fourth layers to improve learning stability. Both networks are trained using PyTorch on an NVIDIA RTX 4090 GPU for 50,000 epochs with the Adam optimizer and a learning rate of $2 \times 10^{-4}$.

\textbf{Training Results:}
The environment CDF model is evaluated on a held-out test set consisting of 50,000 triplets of configurations ($\mathbf{q}$), workspace points ($\mathbf{p}$), and corresponding CDF values. It achieves a Mean Absolute Error (MAE) of 0.041 radians and a Root Mean Square Error (RMSE) of 0.073 radians. Similarly, the SCDF model is evaluated on 10,000 held-out configurations with known self-collision distances, achieving MAE of 0.029 radians and RMSE of 0.054 radians. As shown in Table~\ref{table:inference_time}, the combined inference time for evaluating both the environment and the self-collision CDFs, which together define the neural CDF barrier in \eqref{eq: neural_cdf_barrier}, remains low across different batch sizes. This scalability enables effective integration with sampling-based motion planning algorithms and real-time control using point cloud observations.

\begin{table}[t]
\caption{Inference time (ms) of Neural CDF barrier on xArm for varying batch sizes of configurations ($\mathbf{q}$) and workspace points ($\mathbf{p}$).}
\label{table:inference_time}
\centering
\begin{tabular}{|c|c|c|c|c|c|}
\hline
\diagbox{$\mathbf{q}$}{$\mathbf{p}$} & \textbf{1} & \textbf{10} & \textbf{100} & \textbf{1000} & \textbf{10000} \\
\hline
\textbf{1}   & 0.272 & 0.281 & 0.295 & 0.291 & 0.742 \\
\textbf{10}  & 0.281 & 0.307 & 0.297 & 0.758 & 6.818 \\
\textbf{100} & 0.301 & 0.301 & 0.763 & 7.060 & 70.215 \\
\hline
\end{tabular}
\end{table}

\subsection{Baselines and Parameters}
\label{subsec:baselines}

We compare our approach against the following baselines.
\begin{itemize}
    \item \textbf{Planners:}
    \begin{itemize}
        \item CDF-RRT: RRT with collision checks based on the neural CDF barrier.
        
        \item SDF-RRT \cite{zhu2024efficient}: RRT with collision checks using a neural SDF barrier.
        
        \item SDF-RRT-Connect \cite{rrt_connect_2000}: A variant of SDF-RRT with a bi-directional search strategy.
        
        \item SDF-Lazy-RRT \cite{Hauser_2015_planning}: A variant of SDF-RRT with reduced collision-checking overhead.

        \item MPPI \cite{williams2016aggressive}: A sampling-based MPC method. We compare against the initial “bootstrapping’’ iterations used for planning. 
        
    \end{itemize}
    \item \textbf{Controllers:}
    \begin{itemize}
        \item PD Controller: A proportional-derivative controller $\bar{\bfu}(\bfq, t)$ focused solely on trajectory tracking.
        
        \item PD + CBF-QP \cite{chi2024safe}: A PD controller \(\bar{\bfu}(\bfq, t)\) augmented with a CBF-QP safety filter, where the CBF is constructed using the neural CDF barrier. However, this approach disregards uncertainty in the CDF model or sensor measurements, unlike ours.

        \item MPPI: \cite{williams2016aggressive}: A real-time control baseline that applies MPPI updates at every timestep.
        
    \end{itemize}
\end{itemize}

The baseline planners are implemented using the Open Motion Planning Library (OMPL) \cite{sucan2012the-open-motion-planning-library}, ensuring consistency across all algorithms. All controllers are implemented using Casadi \cite{Andersson2019}, which solves the CBF-QP and our DR-CBF-QP using an interior point optimizer at a rate of $50$ Hz.

\textbf{Evaluation Metrics:} For planners, we evaluate the number of collision checks and path length. For controllers, the evaluation metrics include success rate (percentage of trials where the robot safely reaches the goal) and tracking error (measured as the Fréchet distance between planned and executed paths).

\textbf{Parameters:} Tables~\ref{table:planner_parameters} and~\ref{table: control_parameters} summarize the parameters used for the planners and controllers. For the DR-CBF-QP controller, the Wasserstein radius $r$ and the number of uncertainty samples $N$ are chosen to balance robustness and real-time performance. Following standard practice in data-driven DRO control, we treat $r$ as a tunable robustness parameter. Empirically, values in the range $r \in [0.01, 0.04]$ yield reliable performance: larger radii can lead to overly conservative behavior, whereas very small radii may reduce robustness to perception noise and dynamic obstacles. The sample size is limited by the $50$\,Hz real-time control requirement; using $N = 10$ Monte Carlo dropout samples provides a good tradeoff between robustness and computational efficiency.


\begin{table}[t]
\caption{Planner parameters used for evaluation. \textbf{Goal bias} specifies the probability of sampling configurations directly toward the goal.  \textbf{Step size} defines the maximum extension distance during tree expansion. \textbf{Safety margin} indicates the minimum clearance required for collision checking, and \textbf{collision check resolution} determines the granularity of edge validity checks, expressed as a fraction of the configuration space diagonal. For the 2-DoF planar robot, a resolution of 0.01 corresponds to 0.089 radians, while for the 6-DoF xArm, a resolution of 0.002 corresponds to 0.04 radians.}
\label{table:planner_parameters}
\centering
\resizebox{\linewidth}{!}{%
\begin{tabular}{|l|c|c|c|c|}
\hline
\textbf{Planner} & \textbf{Goal Bias} & \textbf{Step Size} & \textbf{Safety Margin} & \textbf{Col. Check Resolution} \\ \hline
Baselines & 0.1 & 0.1 & 0.05 & 0.01 (2-DoF) / 0.002 (6-DoF) \\ \hline
Bubble-CDF & 0.1 & N/A & 0.05 & N/A \\ \hline
\end{tabular}
}
\end{table}

\begin{table}[ht]
\caption{Control parameters used for evaluation. The PD control gains \( K_P \) and \( K_D \) are positive definite matrices, assumed to be diagonal with identical values along the diagonal. The function $\alpha_h$ is a linear class $\mathcal{K}_{\infty}$ function used in the CBF formulation. Parameters $k$ and $\zeta$ define the governor dynamics in \eqref{eq: governor_dynamics}. The risk tolerance $\epsilon$ and sample size $N$ correspond to the DR-CBF in \eqref{eq: dro_cbf_constraint}, while $r$ represents the Wasserstein radius.}
\label{table: control_parameters}
\centering
\begin{tabular}{|l|c|c|c|c|c|c|c|c|}
\hline
\textbf{Parameter} & $K_P$ & $K_D$& $\alpha_h$ &  $k$ & $\zeta$  & $\epsilon$ & $N$ & $r$ \\ \hline
\textbf{Value}     & 0.8  & 0.1      & 1.0        & 0.2  & 12       & 0.1        & 10  & 0.02 \\ \hline
\end{tabular}
\end{table}

\subsection{2-DoF Planar Robot Simulation}

\begin{figure*}[h]
    \centering
    \subcaptionbox{Initial Configuration\label{fig:3a}}
    {\includegraphics[width=0.24\textwidth]{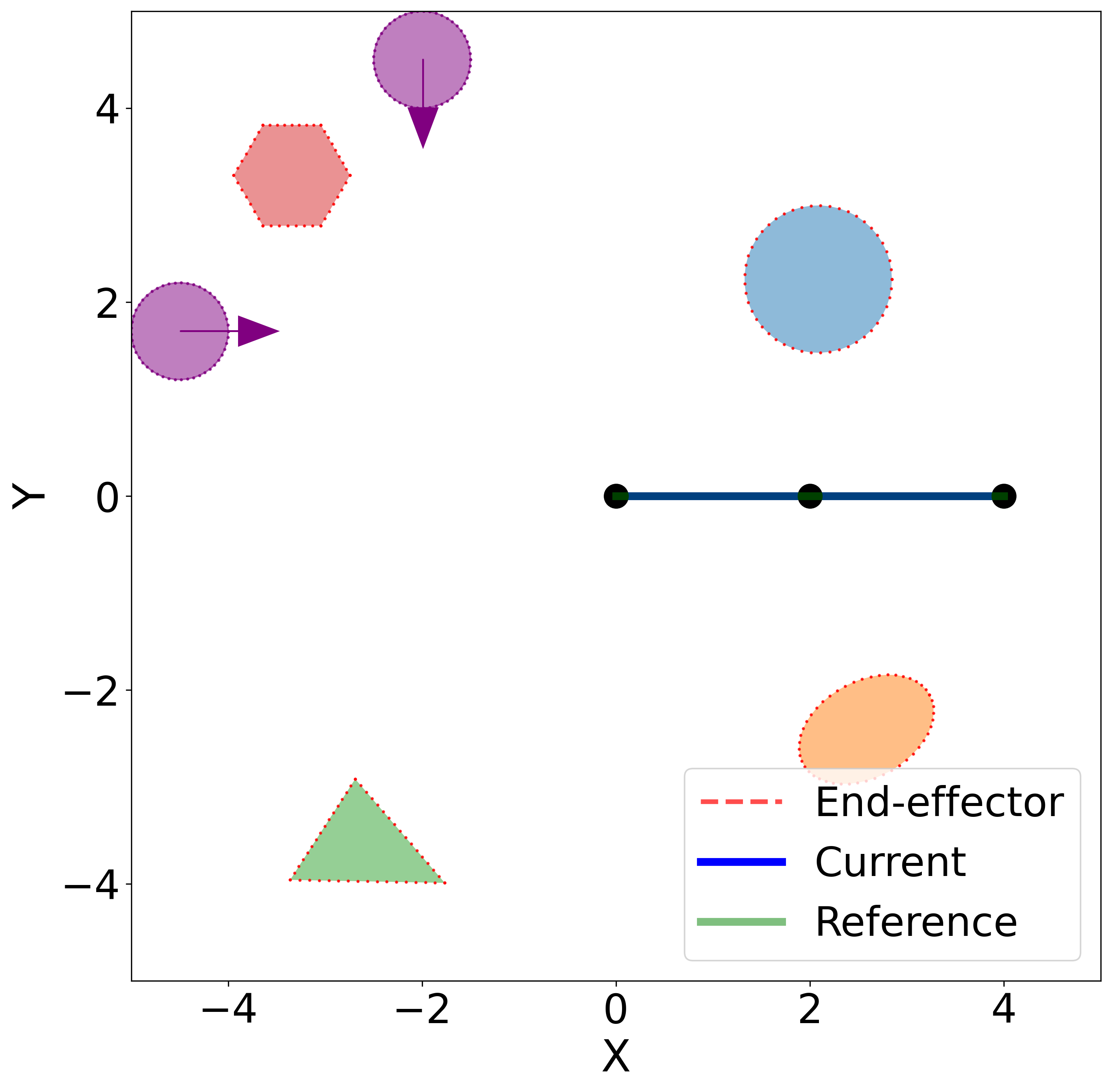}}
    \subcaptionbox{Defensive Maneuver \label{fig:3b}}
    {\includegraphics[width=0.24\textwidth]{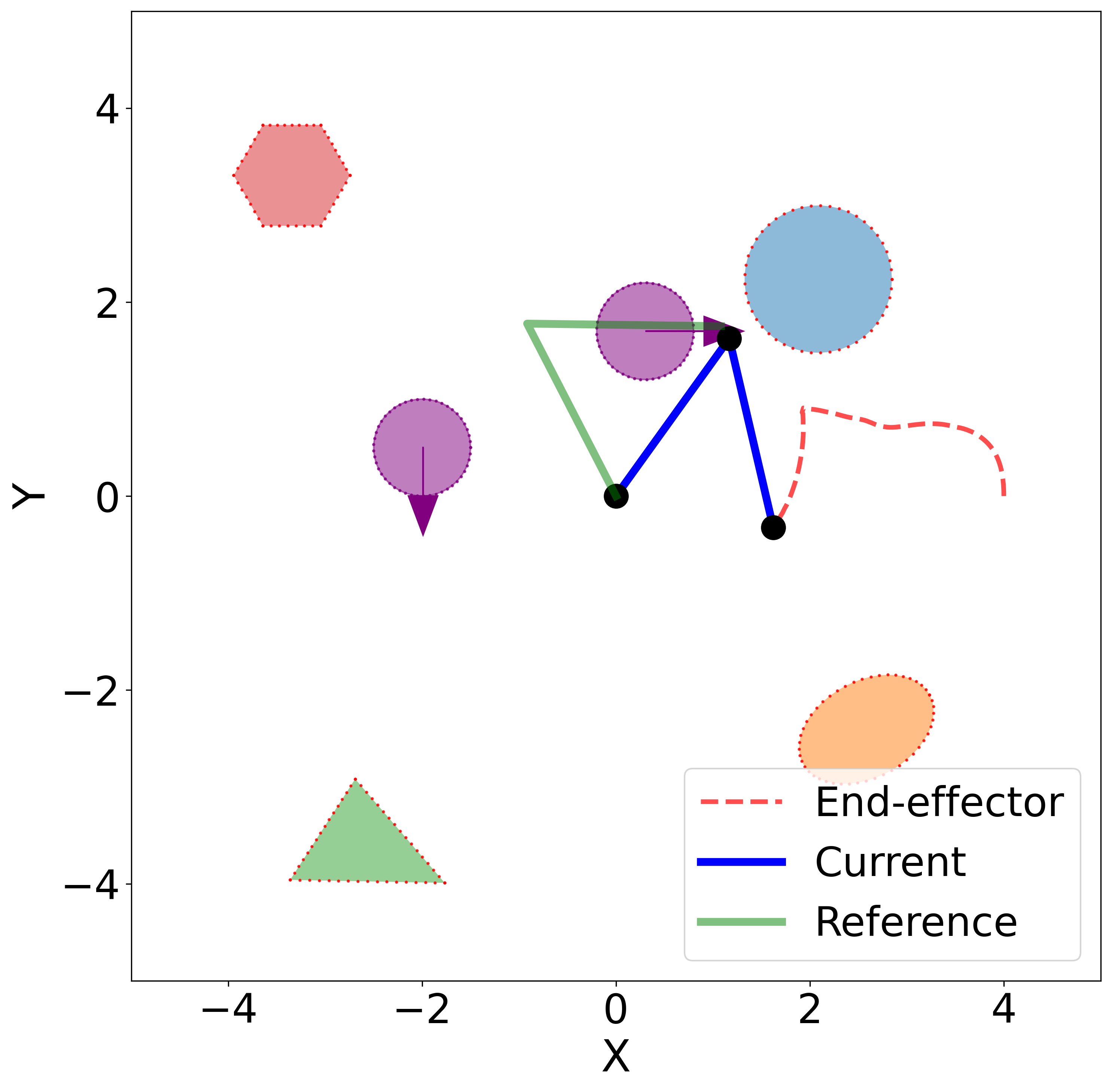}}
    \subcaptionbox{Resuming Tracking \label{fig:3c}}
    {\includegraphics[width=0.24\textwidth]{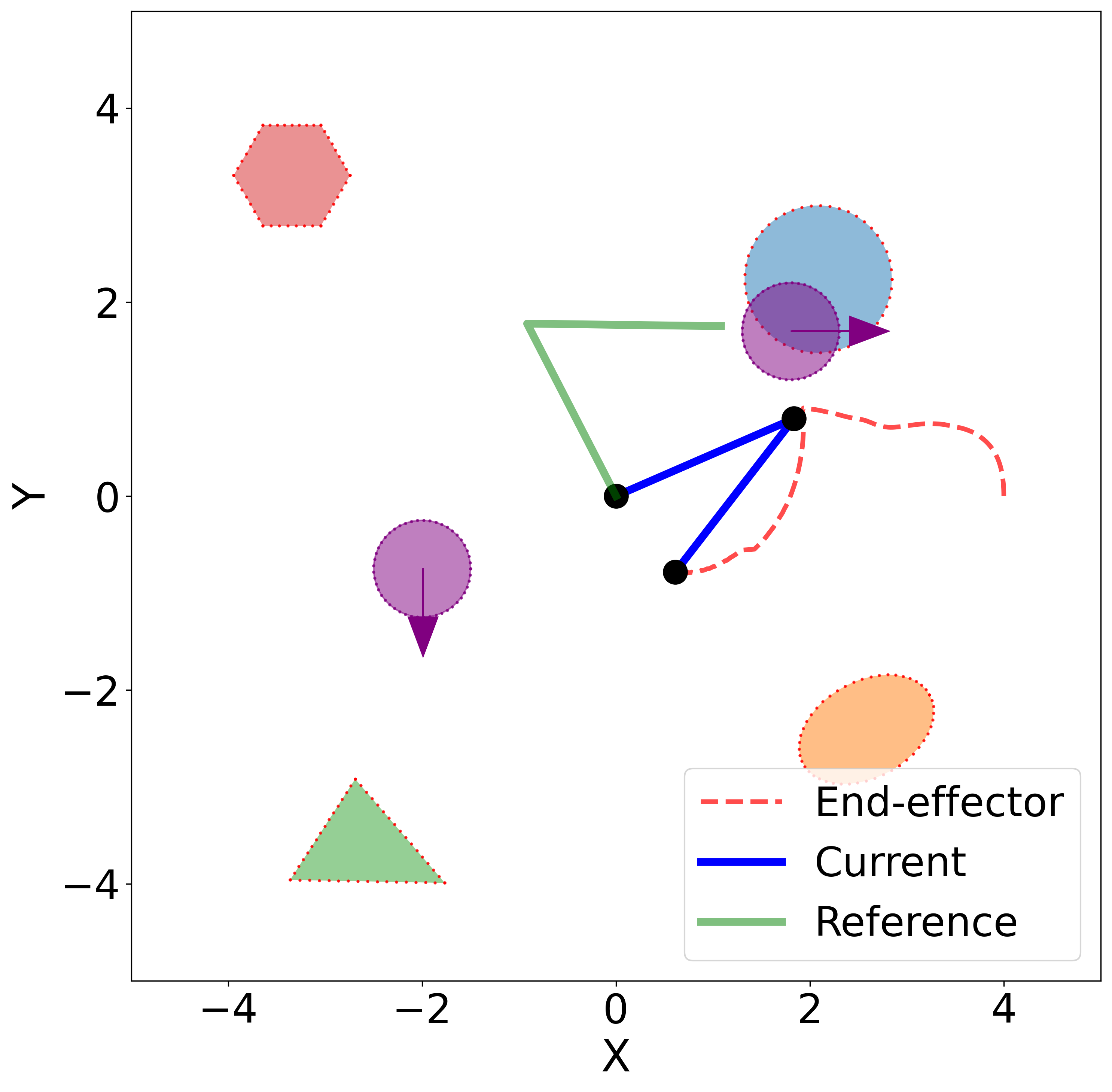}}
    \subcaptionbox{Goal Reached  \label{fig:3d}}
    {\includegraphics[width=0.24\textwidth]{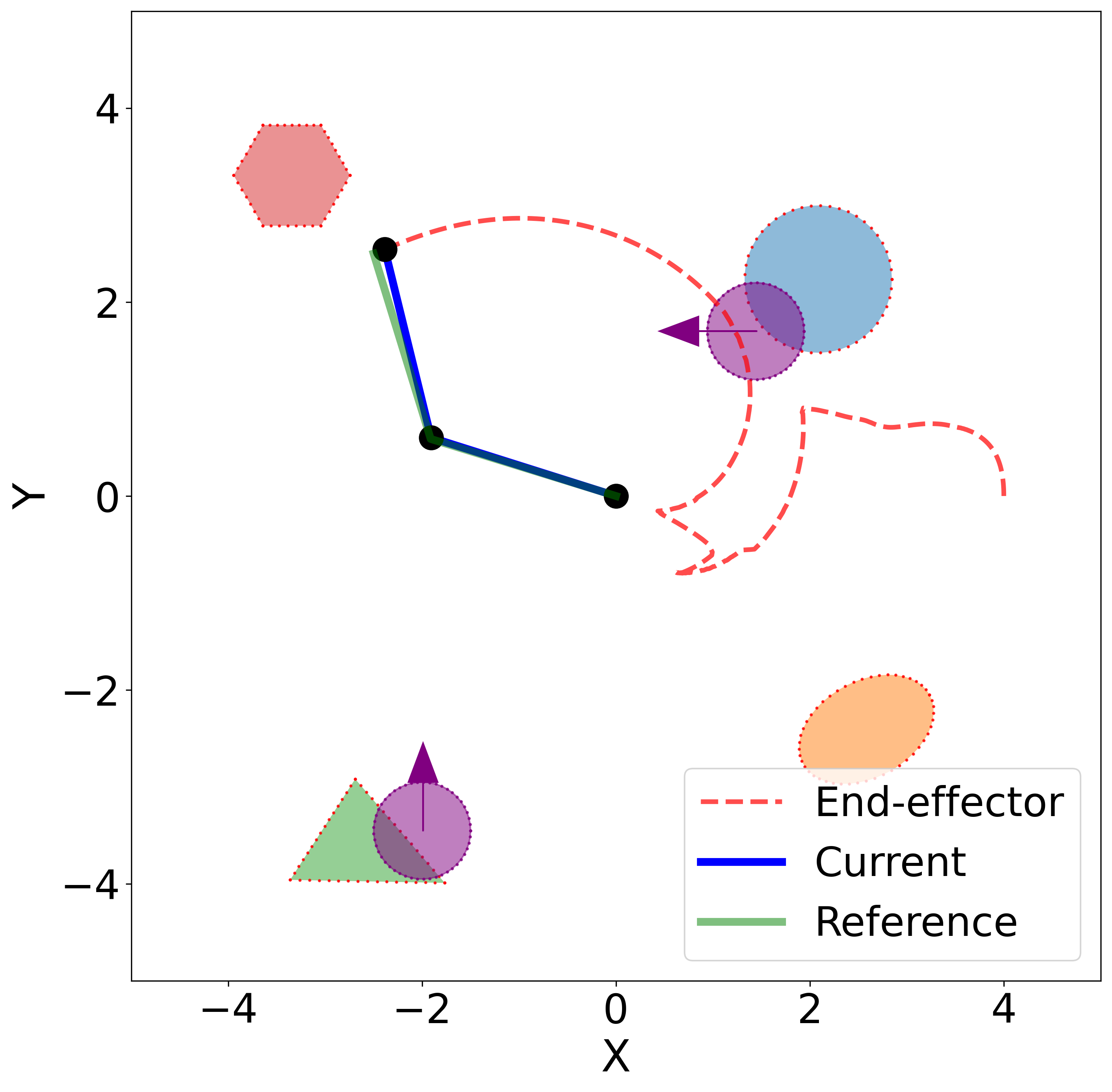}}
    \caption{Snapshots of a 2-link arm navigating a dynamic environment with purple obstacles (velocity directions shown by arrows). The arm is tasked to follow the planned path in Fig.~\ref{fig:2d_example_illustrate}. The arm is shown in blue, and the trajectory of its end-effector is shown in red, and the local reference configuration goal $\gamma(s)$ is shown in green.}
    \label{fig:robot_arm_safe_control}
    \vspace*{-2ex}
\end{figure*}

\begin{table}[t]
\caption{Planning performance comparison on a 2-link arm. }
\label{tab:2dof_results}
\centering
\resizebox{\linewidth}{!}{%
\begin{tabular}{|c|c|c|c|c|}
\hline
\textbf{Planner} & \textbf{Col. Checks} & \textbf{Path Length (rad)} 
& \textbf{Plan Time (s)} & \textbf{Postproc. Time (s)}\\ \hline
CDF-RRT &  $2108.9 \pm 438.7$ & $3.64 \pm 0.69$ 
& $0.08 \pm 0.05$ & $0.22 \pm 0.07$ \\ \hline
SDF-RRT \cite{zhu2024efficient} & $2118.3 \pm 388.9$  & $3.61 \pm 0.69$ 
& $0.08 \pm 0.04$ & $0.23 \pm 0.07$ \\ \hline
SDF-RRT-Connect & $2144.6 \pm 459.7$  & $3.63 \pm 0.74$ 
& $0.03 \pm 0.01$ & $0.22 \pm 0.08$ \\ \hline
SDF-Lazy-RRT & $1759.2 \pm 299.3$  & $3.75 \pm 0.83$ 
& \textbf{$0.02 \pm 0.01$} & $0.23 \pm 0.08$ \\ \hline
MPPI \cite{williams2016aggressive} & $366012.0 \pm 51563.0$ & $5.76 \pm 1.37$ 
& $0.87 \pm 0.15$ & -- \\ \hline
Bubble-CDF & $\boldsymbol{153.8\pm62.2}$ & $3.72\pm0.81$ 
& $0.04 \pm 0.02$ & $\boldsymbol{0.08 \pm 0.02}$ \\ \hline
\end{tabular}}
\vspace{-3ex}
\end{table}

We evaluate our bubble-CDF planner on a 2-DoF planar manipulator with two 2\,m links and a fixed base at the origin.
Joint angles lie in $[-\pi,\pi)$, and the initial configuration $(0,0)$ places the end-effector at $(4,0)$ (Fig.~\ref{fig:2d_example_illustrate}).
All planners operate from this fixed start toward one of two inverse-kinematically valid goal configurations.

\textbf{Setup:} Each of the 500 environments contains four static obstacles of randomized size and position. The end-effector goal is sampled at least 4\,m from the start and guaranteed to be reachable. The robot receives point-cloud observations of obstacle surfaces. All planners must find a collision-free path from the initial configuration to either a valid goal configuration.

\textbf{Results:} Table~\ref{tab:2dof_results} reports performance across all trials. All planners succeed in finding a feasible solution.
The bubble-CDF planner achieves the largest reduction in collision checks, requiring only $153.8 \pm 62.2$ evaluations---over an order of magnitude fewer than the best baseline (Lazy-RRT), which requires at least $10\times$ more collision checks---while maintaining comparable path length.
We report both the planning time (tree expansion/search) and the post-processing time in Table~\ref{tab:2dof_results}. For RRT-based planners, post-processing time refers to shortcutting simplification \cite{rrtrope} and B-spline smoothing, both of which are necessary for practical applications. For the bubble-CDF planner, the post-processing time refers to the convex QCQP solution for trajectory optimization.
Bubble-CDF achieves the lowest overall runtime, with a planning time of $0.04 \pm 0.02$\,s and a trajectory-optimization time of $0.08 \pm 0.02$\,s.
In contrast, the RRT-based baselines require thousands of collision checks and incur similar post-processing costs (approximately $0.23$\,s), even when their planning times are small.
In contrast, MPPI requires far more collision evaluations because it checks collisions along every state of every sampled trajectory in each planning iteration, which results in noticeably slower planning times than graph-based planners.
These results highlight the efficiency of bubble-based configuration-space exploration.

\begin{table}[t]
\caption{Control performance comparison on a 2-link arm. }
\label{tab:2dof_control_results}
\centering
\resizebox{\linewidth}{!}{%
\begin{tabular}{|c|c|c|c|c|}
\hline
\multirow{2}{*}{\textbf{Controller}} & \multicolumn{2}{c|}{\textbf{Static}} & \multicolumn{2}{c|}{\textbf{Dynamic}} \\ \cline{2-5} 
 & \textbf{Success Rate} & \textbf{Tra. Error} & \textbf{Success Rate} & \textbf{Tra. Error} \\ \hline
PD &  $0.874$  & $0.068 \pm 0.017$ &  $0.112$ & $0.062 \pm 0.027$ \\ \hline
PD + CBF-QP \cite{chi2024safe} & $0.982$ & $0.131 \pm 0.049$  & $0.638$  & $0.217 \pm 0.106$  \\ \hline
MPPI \cite{williams2016aggressive} & $\new{1.0}$ & N/A & 0.684 & N/A \\ \hline
PD + DR-CBF-QP & $\boldsymbol{1.0}$ & $0.173 \pm 0.074$ & $\boldsymbol{0.992}$ & $0.394 \pm 0.172$  \\ \hline
\end{tabular}}
\end{table}

Next, we present our evaluation of the DR-CBF-QP formulation for safe control synthesis on a planar robot. 

\textbf{Setup:} Controllers are tested in both static (Fig.~\ref{fig:2d_example_illustrate}) and dynamic environments (Fig.~\ref{fig:robot_arm_safe_control}) using point-cloud observations. For dynamic settings, obstacle velocities are sampled as $v \sim \mathcal{N}(0.5,0.1)$\,m/s, while the robot assumes a nominal velocity of $0.5$\,m/s for control synthesis. Each controller is evaluated over 500 random trials for both environment types. The task is to track the bubble-CDF trajectory $\gamma$ while avoiding collisions.

\textbf{Results:} Table~\ref{tab:2dof_control_results} summarizes the control performance. DR-CBF-QP attains perfect success in static environments and $99.2\%$ success in dynamic settings, outperforming PD, standard CBF-QP, and MPPI.

The PD controller, designed only for tracking, performs poorly in both scenarios.
While CBF-QP performs well in static environments, its success rate drops sharply under dynamic conditions ($63.8\%$), as it does not account for obstacle-velocity uncertainty. MPPI achieves slightly higher success rates compared to CBF QP, as it optimizes over a receding horizon and can anticipate future states. However, it still does not account for uncertainty, which limits its performance in dynamic environments. 

DR-CBF-QP exhibits higher tracking error than the baselines, reflecting a deliberate prioritization of safety: the controller modifies the trajectory as necessary to preserve collision margins, particularly under dynamic uncertainty.
Qualitative results in Fig.~\ref{fig:robot_arm_safe_control} show the robot executing defensive maneuvers (Fig.~\ref{fig:3b}) to avoid moving obstacles before rejoining the nominal plan (Fig.~\ref{fig:3c}) and reaching the goal (Fig.~\ref{fig:3d}).

\subsection{6-DoF xArm Robot Simulation}
We evaluate our planning and control methods on a 6-DoF xArm manipulator simulated in PyBullet \cite{coumans2021_pybullet}, considering both static and dynamic obstacles.

\begin{figure*}[h]
    \centering
    \subcaptionbox{Initial Configuration\label{fig:4a}}
    {\includegraphics[width=0.25\linewidth, trim={20cm 0cm 15cm 15cm},clip]{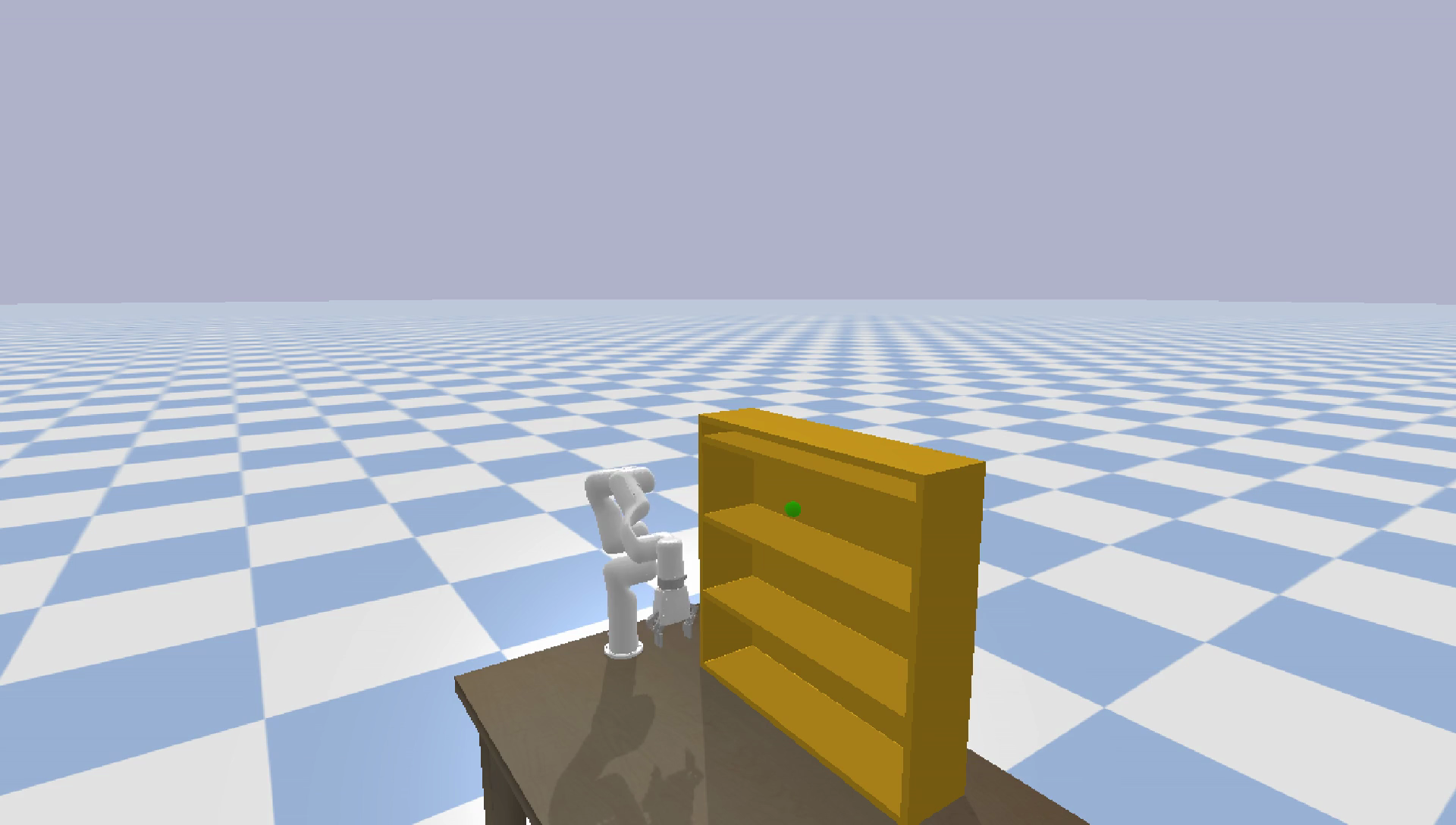}}%
    \subcaptionbox{ Planned Configuration 1\label{fig:4b}}
    {\includegraphics[width=0.25\linewidth, trim={20cm 0cm 15cm 15cm},clip]{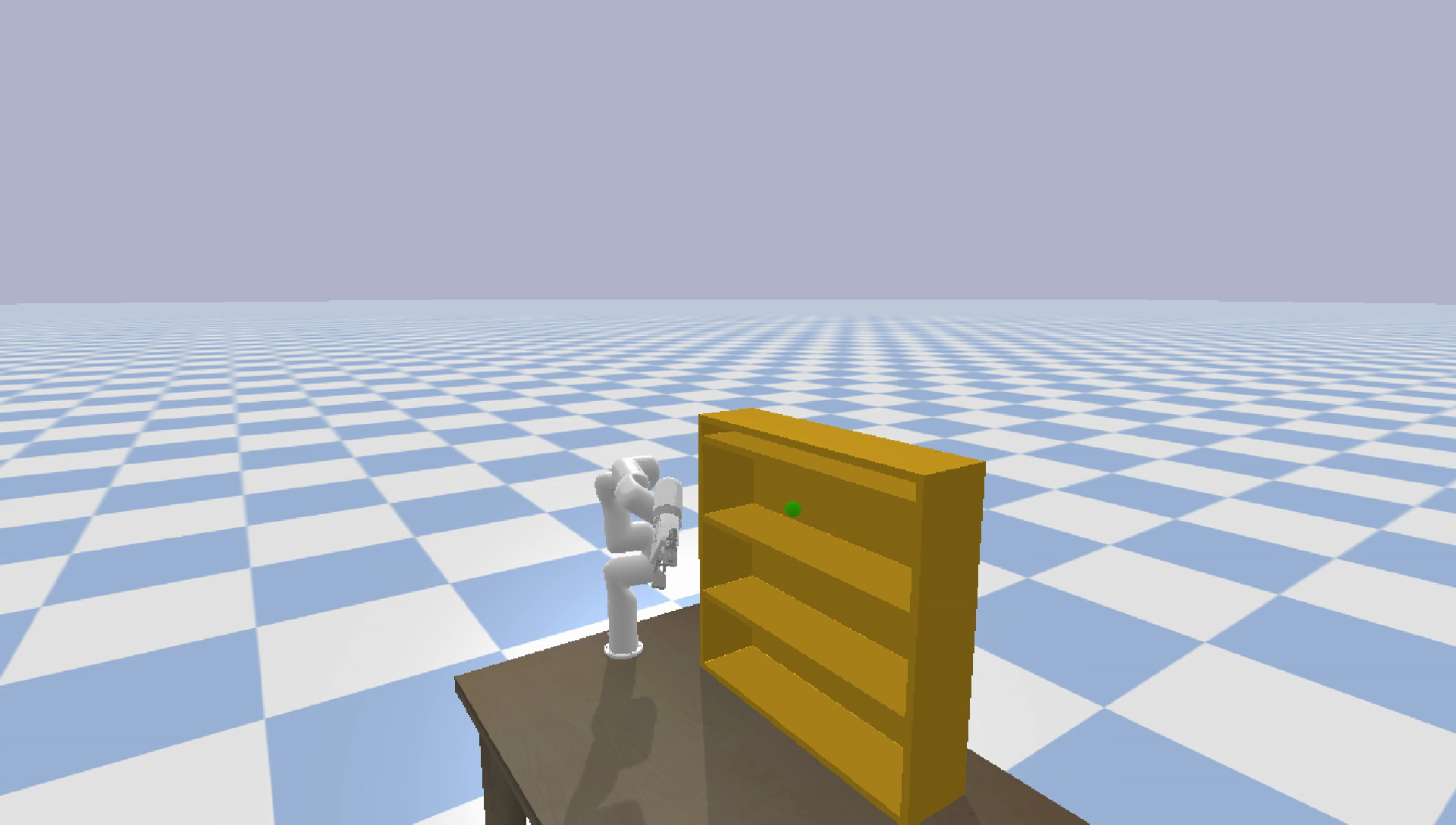}}%
    \subcaptionbox{Planned Configuration 2\label{fig:4c}}
    {\includegraphics[width=0.25\linewidth, trim={20cm 0cm 15cm 15cm},clip]{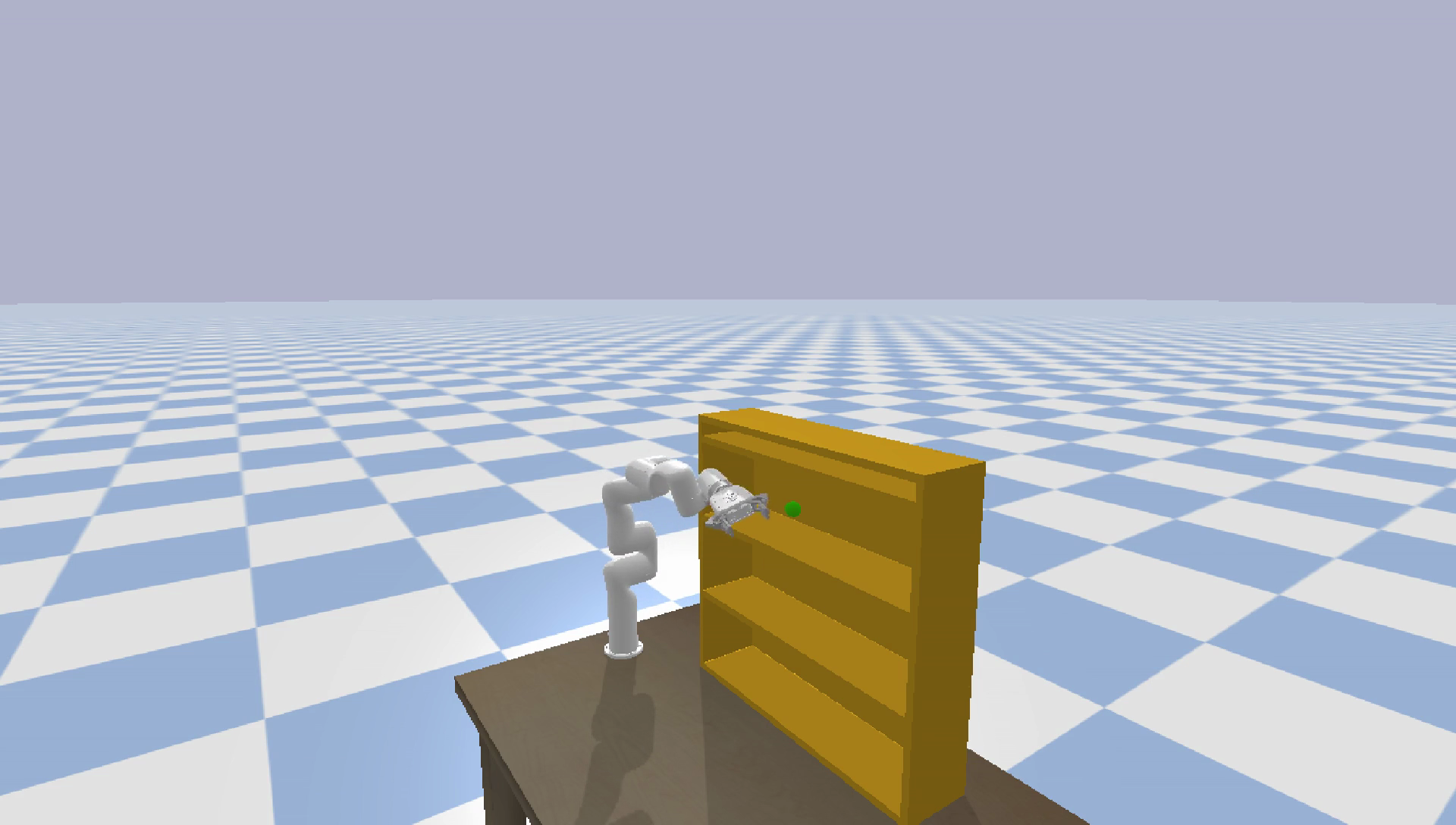}}%
    \subcaptionbox{Goal Configuration \label{fig:4d}}
    {\includegraphics[width=0.25\linewidth, trim={20cm 0cm 15cm 15cm},clip]{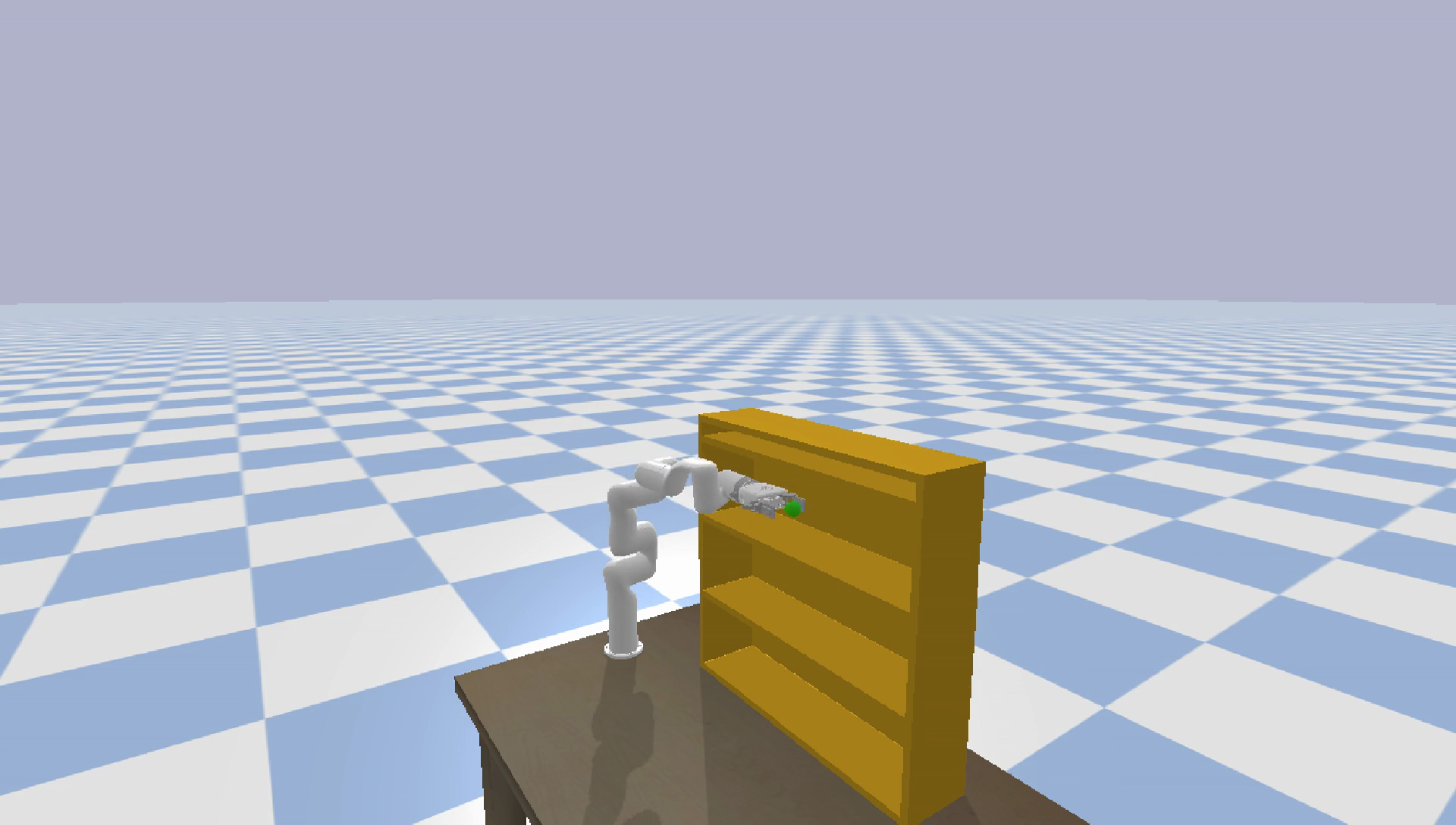}}
    \caption{Bubble-CDF planning for a 6-DoF xArm robot in a static environment, targeting an end-effector goal represented by a green sphere. (a) The initial configuration of the xArm. (b, c) Intermediate configurations illustrating the planned path as the robot avoids obstacles while progressing toward the goal. (d) The final goal configuration reached by the robot.}
    \label{fig: pybullet_planning}
\end{figure*}

\textbf{Setup:} The environment consists of a 6-DoF xArm mounted on a table adjacent to a shelf (Fig.~\ref{fig: pybullet_planning}). A depth camera provides point-cloud observations of surrounding obstacles. We perform 50 randomized trials by varying the shelf position and sampling different end-effector goals. For each trial, five feasible goal configurations are generated using inverse kinematics, and motion planning is executed in the static environment. To evaluate controller performance, the best bubble-CDF planned path for each trial is selected for tracking. The dynamic environment includes three moving obstacles with known positions and velocities, sampled from $\mathcal{N}(0.2, 0.05)$ m/s.


\begin{table}[t]
\caption{Planning performance comparison on a 6-DoF xArm robot in PyBullet Simulation.} 
\label{tab: xarm_planning_results}
\centering
\resizebox{\linewidth}{!}{%
\begin{tabular}{|c|c|c|c|c|}
\hline
\textbf{Planner} & \textbf{Col. Checks} & \textbf{Path Length} 
& \textbf{Plan Time (s)} & \textbf{Postproc. Time (s)}\\ \hline
CDF-RRT &  $3415.4 \pm 803.6$ & $3.11 \pm 0.46$ 
& $0.32 \pm 0.05$ & $0.47 \pm 0.13$ \\ \hline
SDF-RRT \cite{zhu2024efficient} & $3251.8 \pm 782.4$  & $3.07 \pm 0.49$ 
& $0.33 \pm 0.08$ & $0.50 \pm 0.11$ \\ \hline
SDF-RRT-Connect  & $3437.4 \pm 723.2$  & $3.11 \pm 0.43$ 
& $0.04 \pm 0.01$ & $0.55 \pm 0.13$ \\ \hline
SDF-Lazy-RRT  & $2741.1 \pm 737.9$  & $3.15 \pm 0.53$ 
& $0.04 \pm 0.01$ & $0.48 \pm 0.13$ \\ \hline
MPPI \cite{williams2016aggressive} & $1331893.2 \pm 500936.1$ & $6.83 \pm 1.22$ 
& $4.97 \pm 0.83$ & -- \\ \hline
Bubble-CDF & $\boldsymbol{278.5\pm80.7}$ & $3.04\pm0.46$ 
& $0.11 \pm 0.03$ & $\boldsymbol{0.08 \pm 0.03}$  \\ \hline
\end{tabular}}
\end{table}

\begin{figure*}[h]
    \centering
    \subcaptionbox{Start of Execution\label{fig:5a}}
    {\includegraphics[width=0.25\linewidth, trim={20cm 0cm 15cm 15cm},clip]{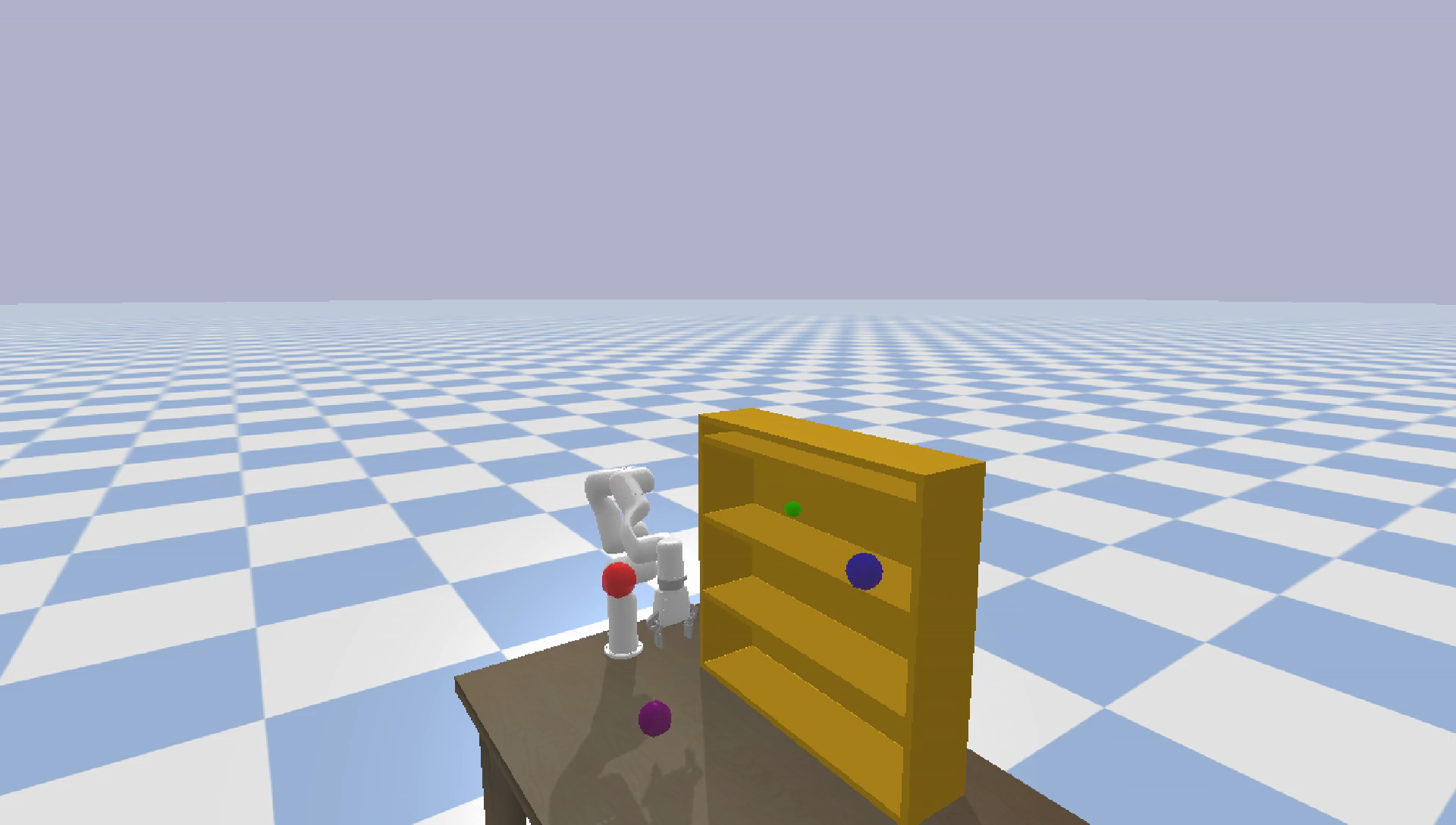}}%
    \subcaptionbox{Dynamic Obstacle Approaching\label{fig:5b}}
    {\includegraphics[width=0.25\linewidth, trim={20cm 0cm 15cm 15cm},clip]{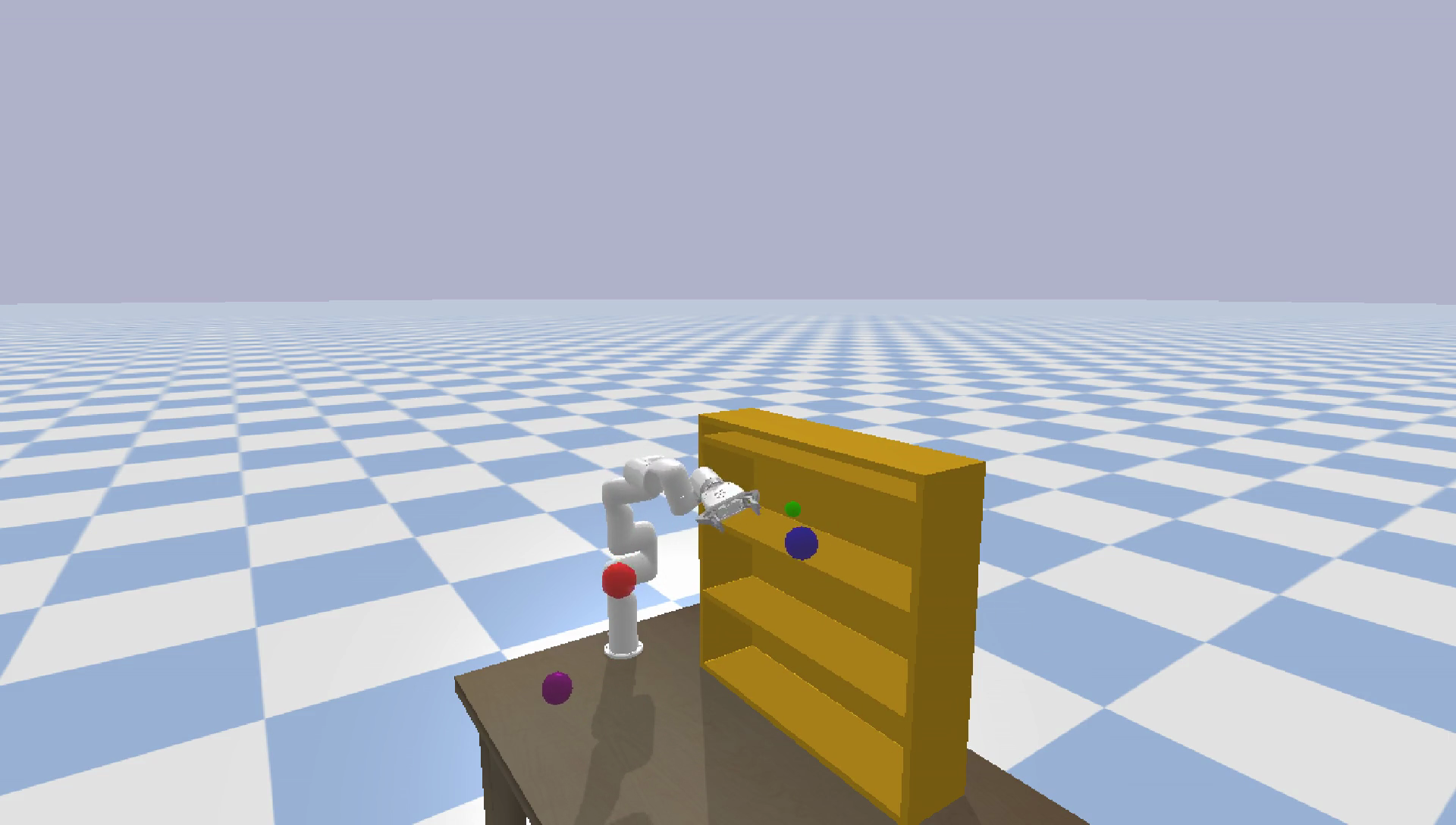}}%
    \subcaptionbox{Defensive Maneuver\label{fig:5c}}
    {\includegraphics[width=0.25\linewidth, trim={20cm 0cm 15cm 15cm},clip]{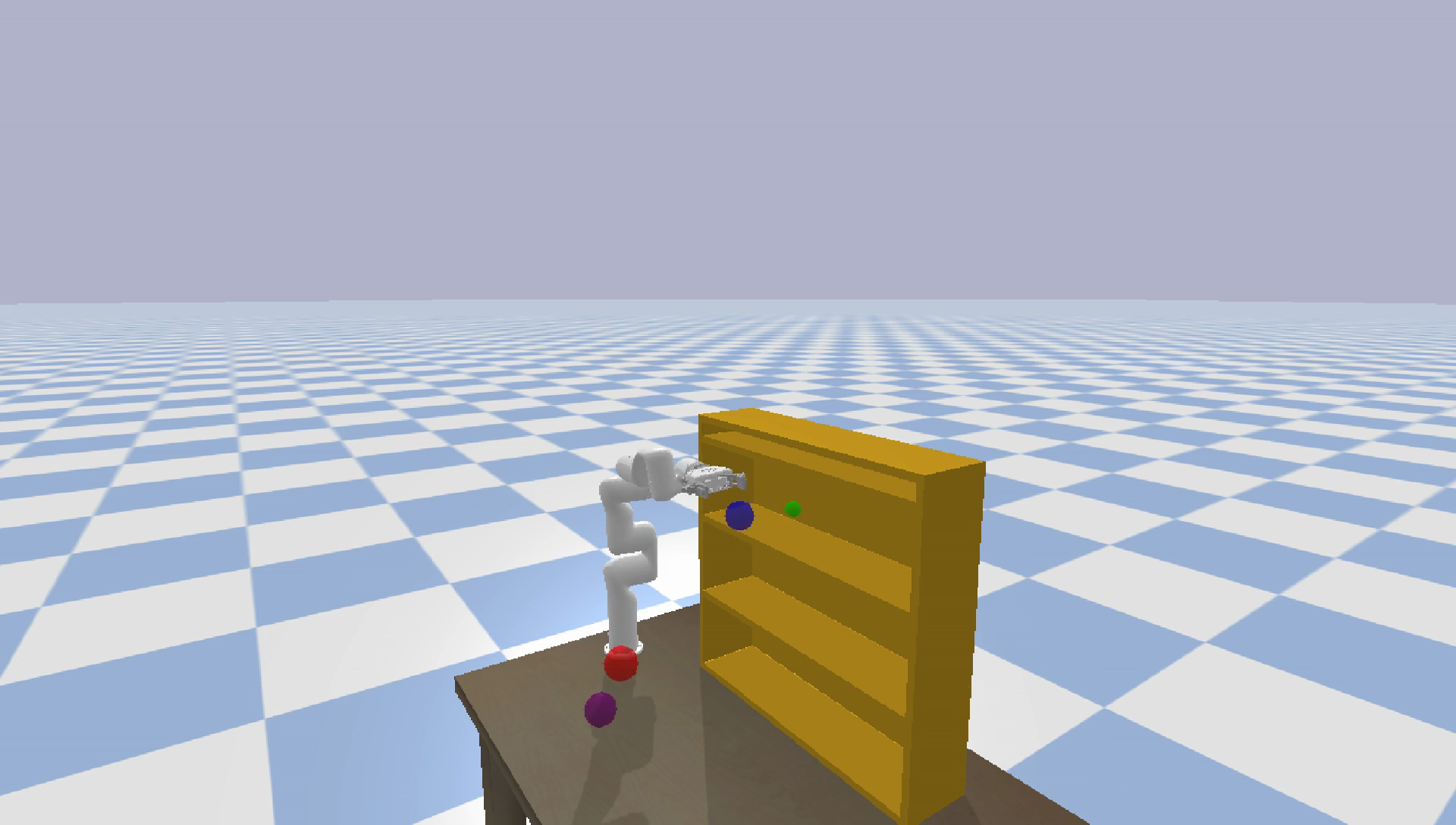}}%
    \subcaptionbox{Goal Reached\label{fig:5d}}
    {\includegraphics[width=0.25\linewidth, trim={20cm 0cm 15cm 15cm},clip]{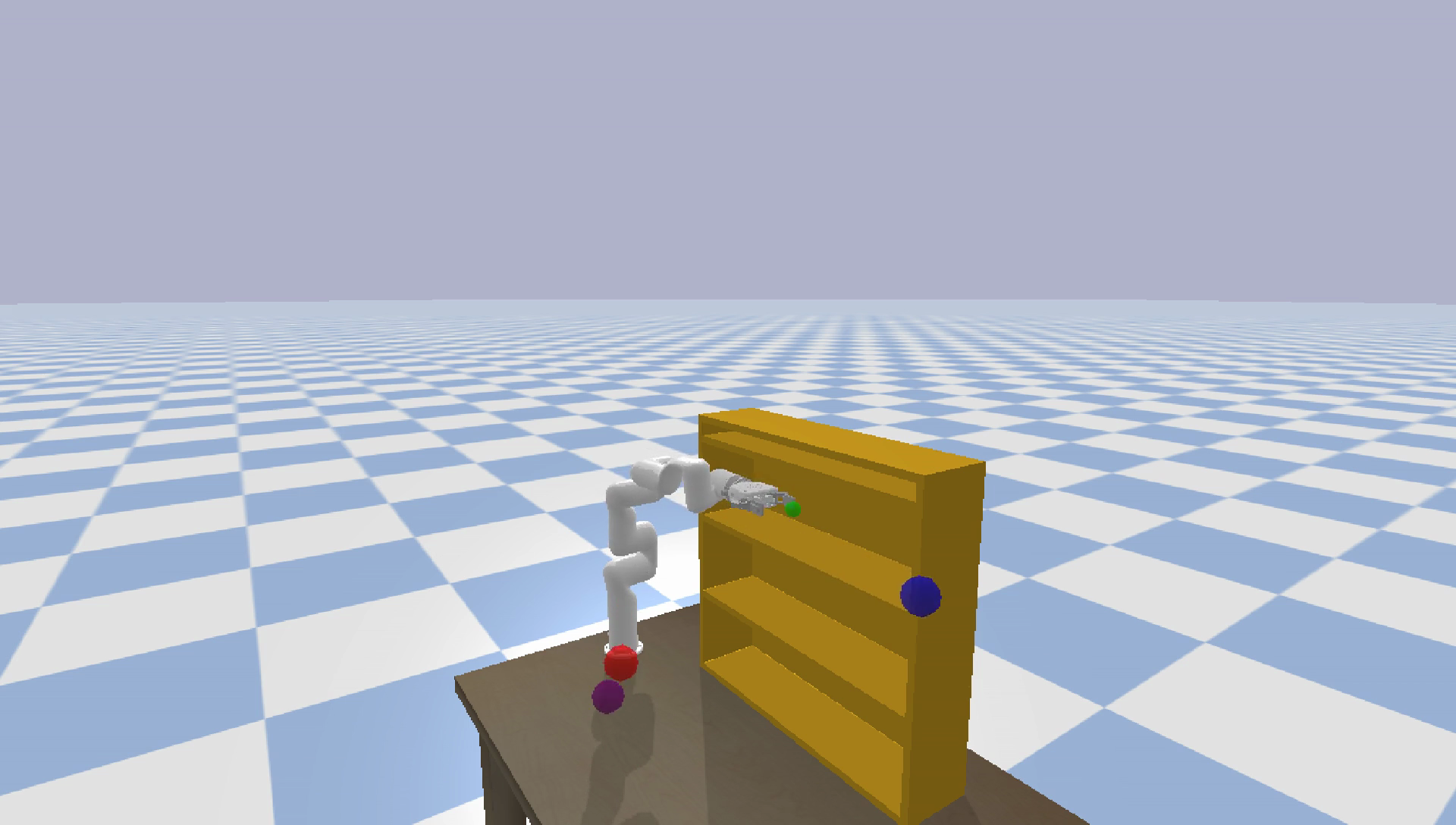}}
    \caption{Snapshots of safe control execution on a 6-DoF xArm robot in an environment with dynamic obstacles. (a) The control execution begins at the initial configuration. (b) A dynamic obstacle (blue) approaches the robot from the right. (c) The robot executes a defensive maneuver, moving upward to avoid the obstacle. (d) The robot successfully resumes tracking and reaches the goal configuration.}
    \label{fig: pybullet_control}
\end{figure*}

\textbf{Results and Discussion:} Table~\ref{tab: xarm_planning_results} reports the planning performance across all 50 trials. As in the 2-DoF benchmark, the bubble-CDF planner achieves a dramatic reduction in collision checks ($278.5 \pm 80.7$) while maintaining path lengths comparable to all baselines.
We report both the planning time and the trajectory post-processing time. The Bubble-CDF planner achieves the lowest overall runtime.
Notably, the trajectory optimization step in Sec.~\ref{subsec:path_optimization} is a convex QCQP and does not require additional collision checks, since safety is enforced by constraining Bézier control points to lie inside the bubbles.
This highlights the efficiency of the bubble-CDF planner in higher-dimensional configuration spaces. Figure~\ref{fig: pybullet_planning} shows a path planned with bubble-CDF for a representative static-environment trial.

Table~\ref{tab: xarm_sim_control_results} summarizes controller performance. The baseline PD controller achieves $68\%$ success in static environments but only $18\%$ in dynamic scenarios, reflecting its lack of collision-awareness. Adding a CBF-QP safety filter increases static-environment success to $88\%$ but remains limited in dynamic settings ($70\%$), as standard CBF-QP does not account for uncertainty in obstacle motion or CDF estimation.

MPPI is run with 300 samples and a horizon of 50. Thanks to its receding horizon formulation, MPPI achieves a higher success rate than the standard CBF QP controller, especially in dynamic environments. 

Our PD + DR-CBF-QP controller achieves $100\%$ success in both static and dynamic environments. While tracking error is higher than the baselines, this reflects its safety-first behavior: the controller adaptively modifies the trajectory to maintain collision margins under dynamic uncertainty, even when deviating from the nominal plan.

Fig.~\ref{fig: pybullet_control} illustrates the execution of a planned path by the 6-DoF xArm robot in a dynamic environment. The robot begins at the initial configuration (Fig.~\ref{fig:5a}) and dynamically reacts to an approaching obstacle (blue) as shown in Fig.~\ref{fig:5b}. By executing a defensive maneuver (Fig.~\ref{fig:5c}), the robot moves upward to maintain safety and avoid a collision. After the obstacle passes, the robot resumes trajectory tracking and successfully reaches the goal configuration (Fig.~\ref{fig:5d}). 

\begin{table}[t]
\caption{Control performance comparison on a 6-DoF xArm robot in PyBullet simulation.}
\label{tab: xarm_sim_control_results}
\centering
\resizebox{\linewidth}{!}{%
\begin{tabular}{|c|c|c|c|c|}
\hline
\multirow{2}{*}{\textbf{Controller}} & \multicolumn{2}{c|}{\textbf{Static}} & \multicolumn{2}{c|}{\textbf{Dynamic}} \\ \cline{2-5} 
 & \textbf{Success Rate}  & \textbf{Tra. Error} & \textbf{Success Rate}  &\textbf{Tra. Error} \\ \hline
PD &  $0.68$  & $0.053 \pm 0.014$ &  $0.18$ &  $0.051 \pm 0.013 $ \\ \hline
PD + CBF-QP \cite{chi2024safe} & $0.88$  & $0.115 \pm 0.046$  & $0.70$ & $0.327 \pm 0.104$  \\ \hline
MPPI \cite{williams2016aggressive}  & $0.9$ & N/A & $0.86$ & N/A \\ \hline
PD + DR-CBF-QP & $1.0$ &  $0.143 \pm 0.055$ & $1.0$  &$0.516 \pm 0.221$  \\ \hline
\end{tabular}}
\end{table}

\subsection{6-DoF xArm Robot Experiments}

\begin{table}[t]
\caption{Planning performance comparison on a real 6-DoF xArm robot.}
\label{tab: real_xarm_planning_results}
\centering
\resizebox{\linewidth}{!}{%
\begin{tabular}{|c|c|c|c|}
\hline
\textbf{Planner} & \textbf{Col. Checks} & \textbf{Path Length} & \textbf{Time (s)} \\ \hline
CDF-RRT &  $5617.4 \pm 1333.6$ & $3.71 \pm 0.71$ & $1.02 \pm 0.33$ \\ \hline
SDF-RRT \cite{zhu2024efficient} & $5911.9 \pm 1487.7$  & $3.72 \pm 0.68$ & $1.04 \pm 0.33$  \\ \hline
SDF-RRT-Connect  & $5801.5 \pm 1265.4$  & $3.75 \pm 0.63$ & $1.05 \pm 0.36$ \\ \hline
SDF-Lazy-RRT  & $4765.8 \pm 1331.4$  & $3.84 \pm 0.69$ & $0.96 \pm 0.32$ \\ \hline
Bubble-CDF & $\boldsymbol{634.8\pm225.1}$ & $3.81\pm0.66$ & $\boldsymbol{0.22 \pm 0.08}$ \\ \hline
\end{tabular}}
\vspace{-3ex}
\end{table}

\begin{figure*}[h]
    \centering
    \subcaptionbox{Planned configurations\label{fig:6a}}
    {\includegraphics[width=0.24\textwidth]{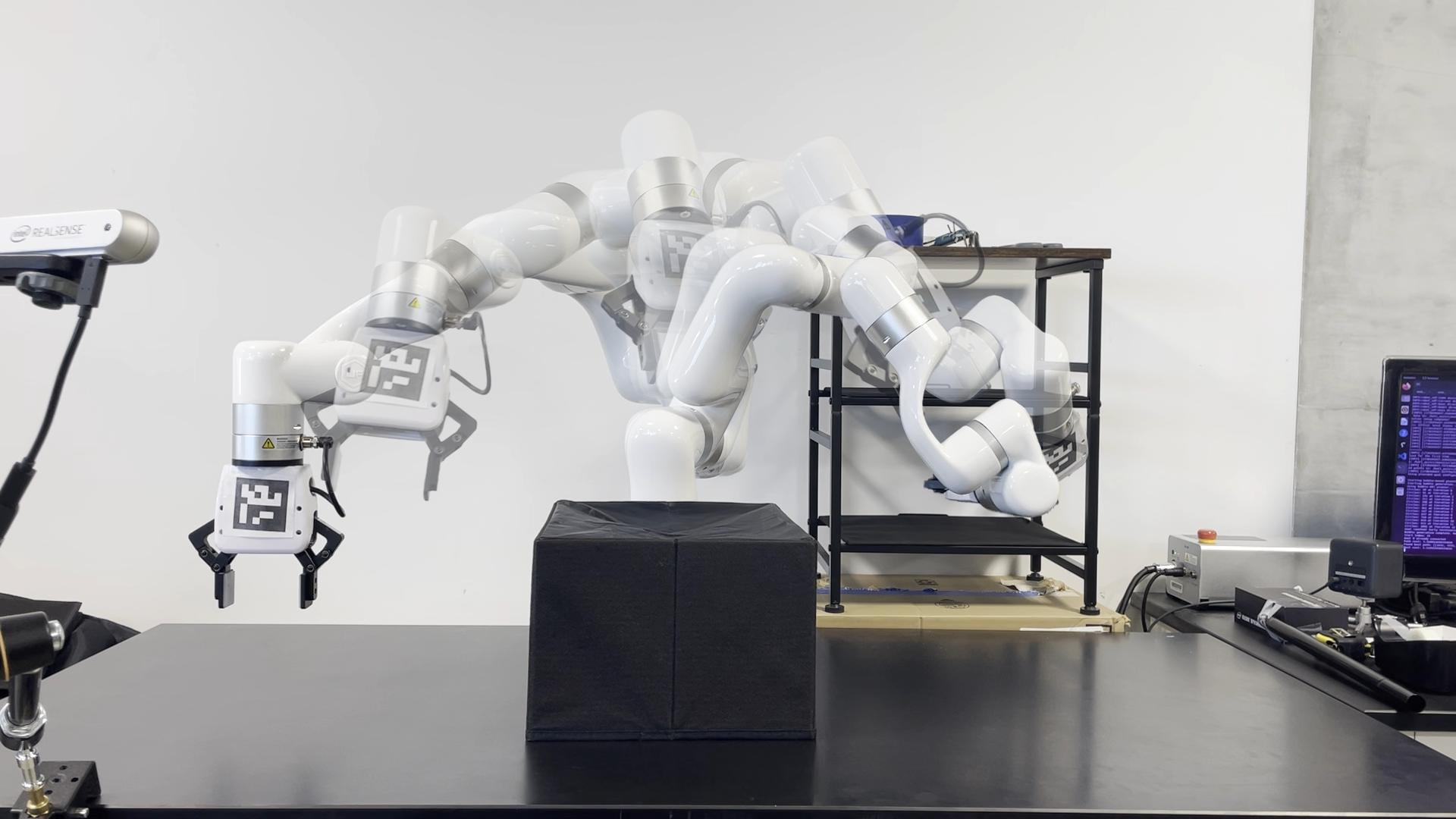}}
    \subcaptionbox{Response to dynamic obstacle\label{fig:6b}}
    {\includegraphics[width=0.24\textwidth]{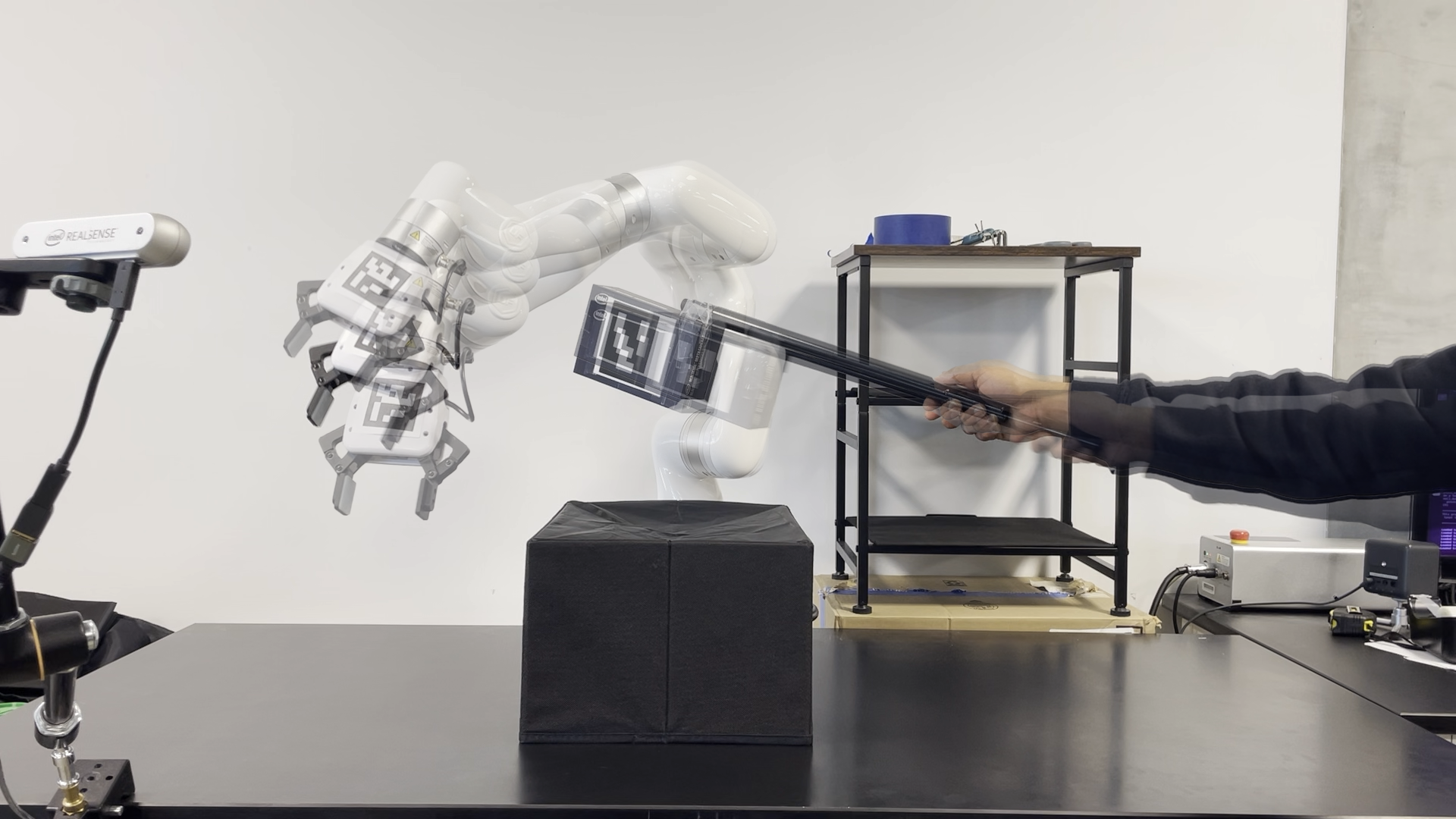}}
    \subcaptionbox{Response to dynamic obstacle\label{fig:6c}}
    {\includegraphics[width=0.24\textwidth]{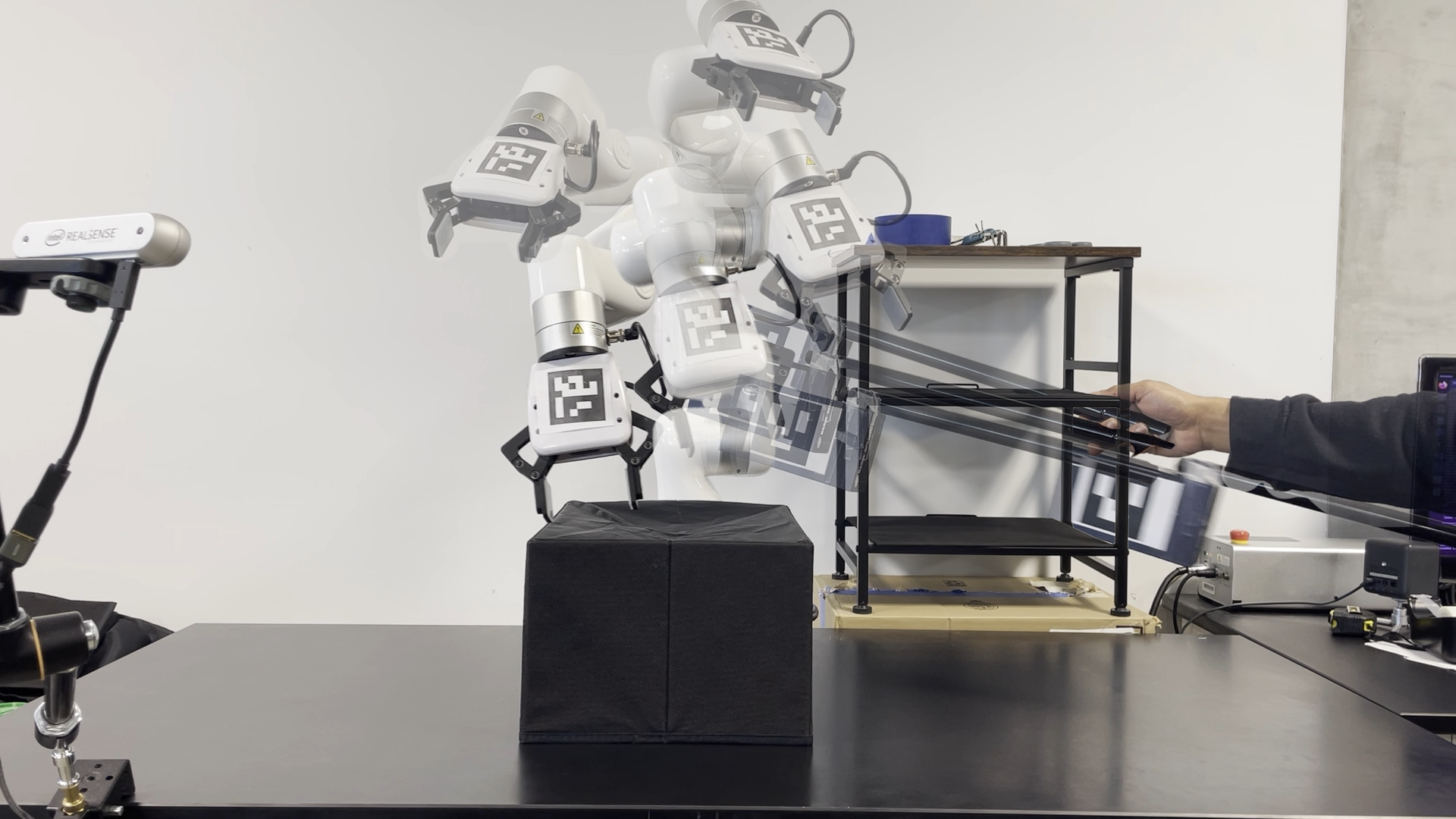}}
    \subcaptionbox{Response to dynamic obstacle\label{fig:6d}}
    {\includegraphics[width=0.24\textwidth]{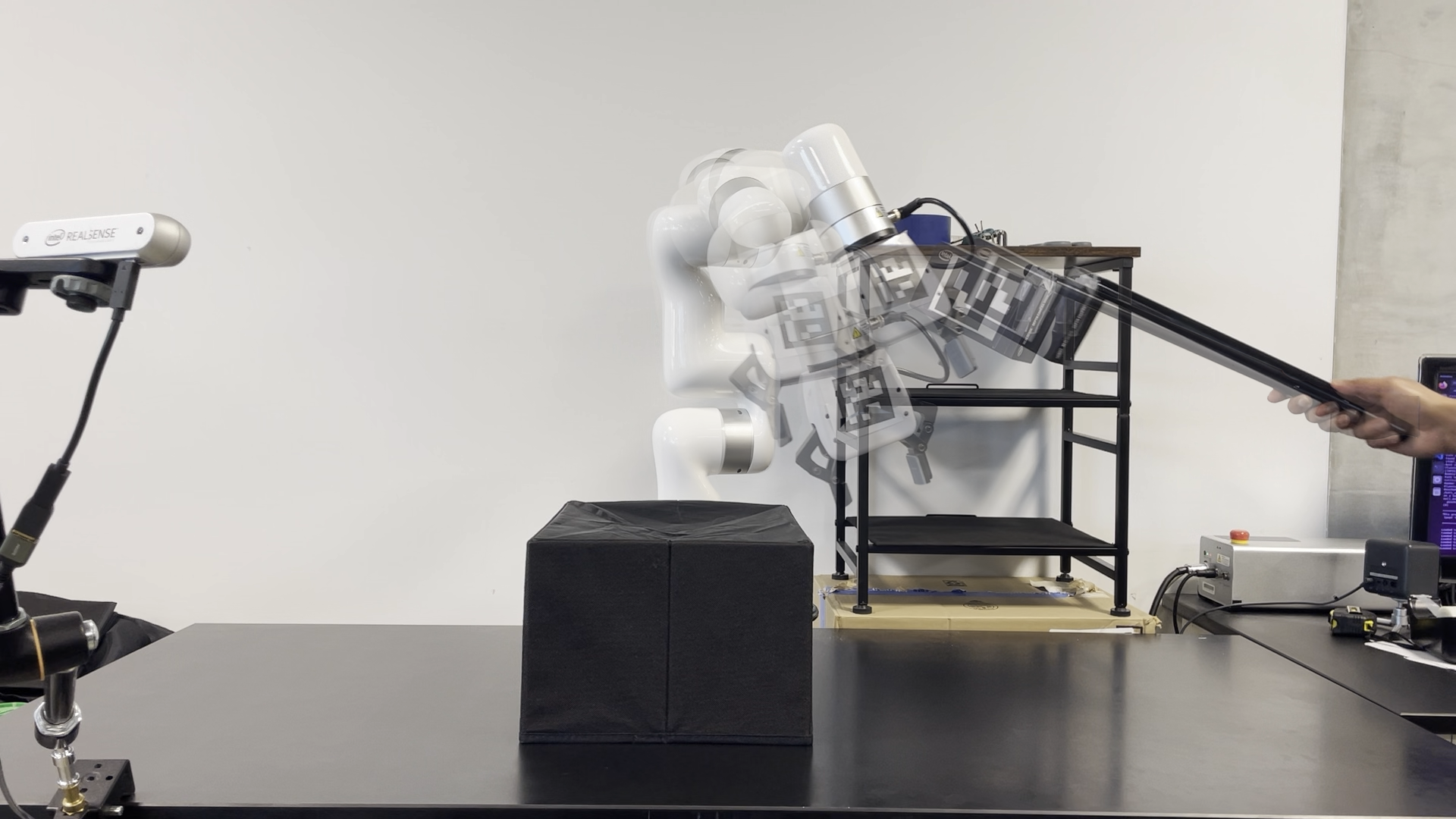}} \\
    \subcaptionbox{Planned configurations\label{fig:7a}}
    {\includegraphics[width=0.24\textwidth]{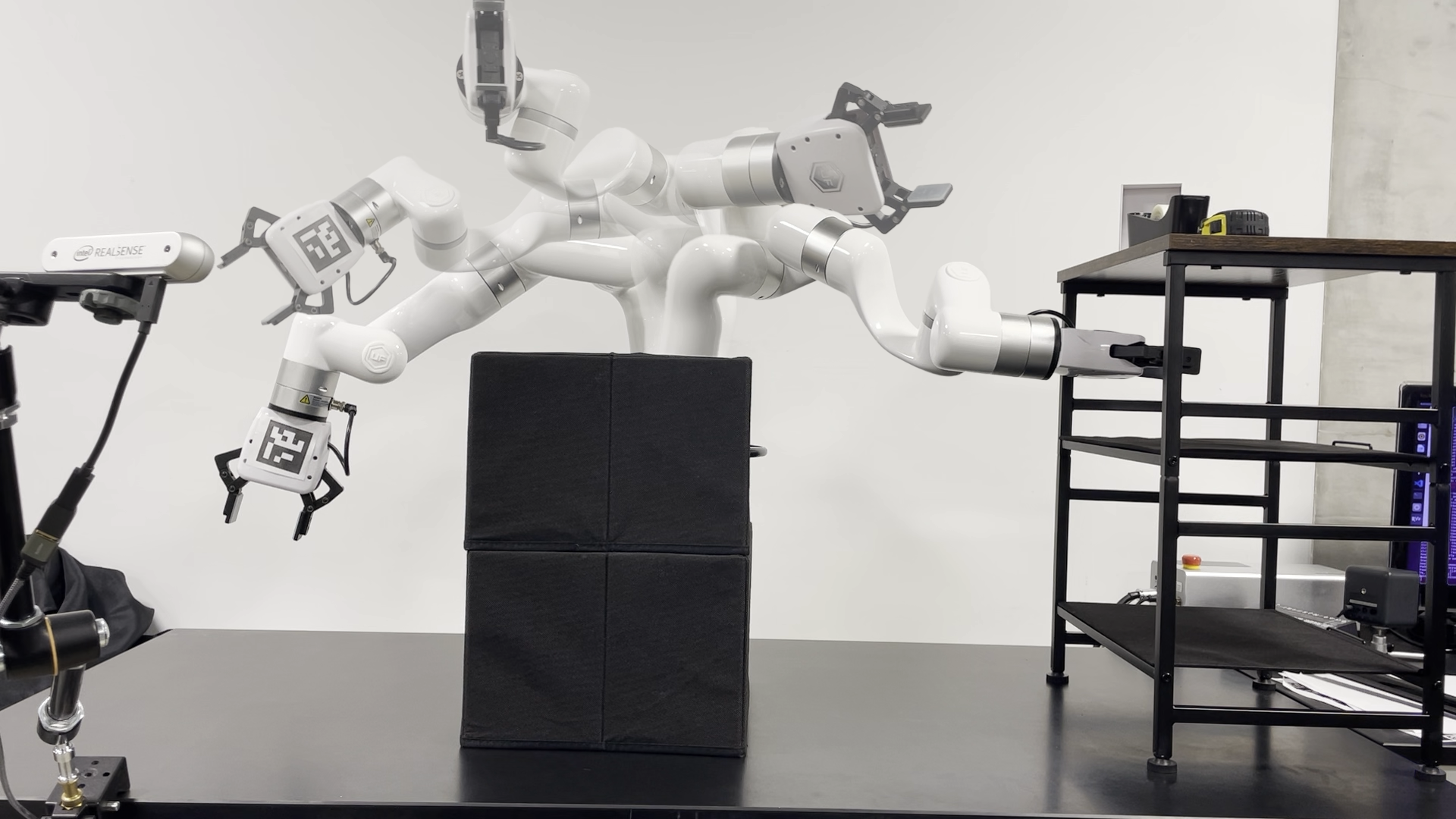}}
    \subcaptionbox{Response to dynamic obstacle \label{fig:7b}}
    {\includegraphics[width=0.24\textwidth]{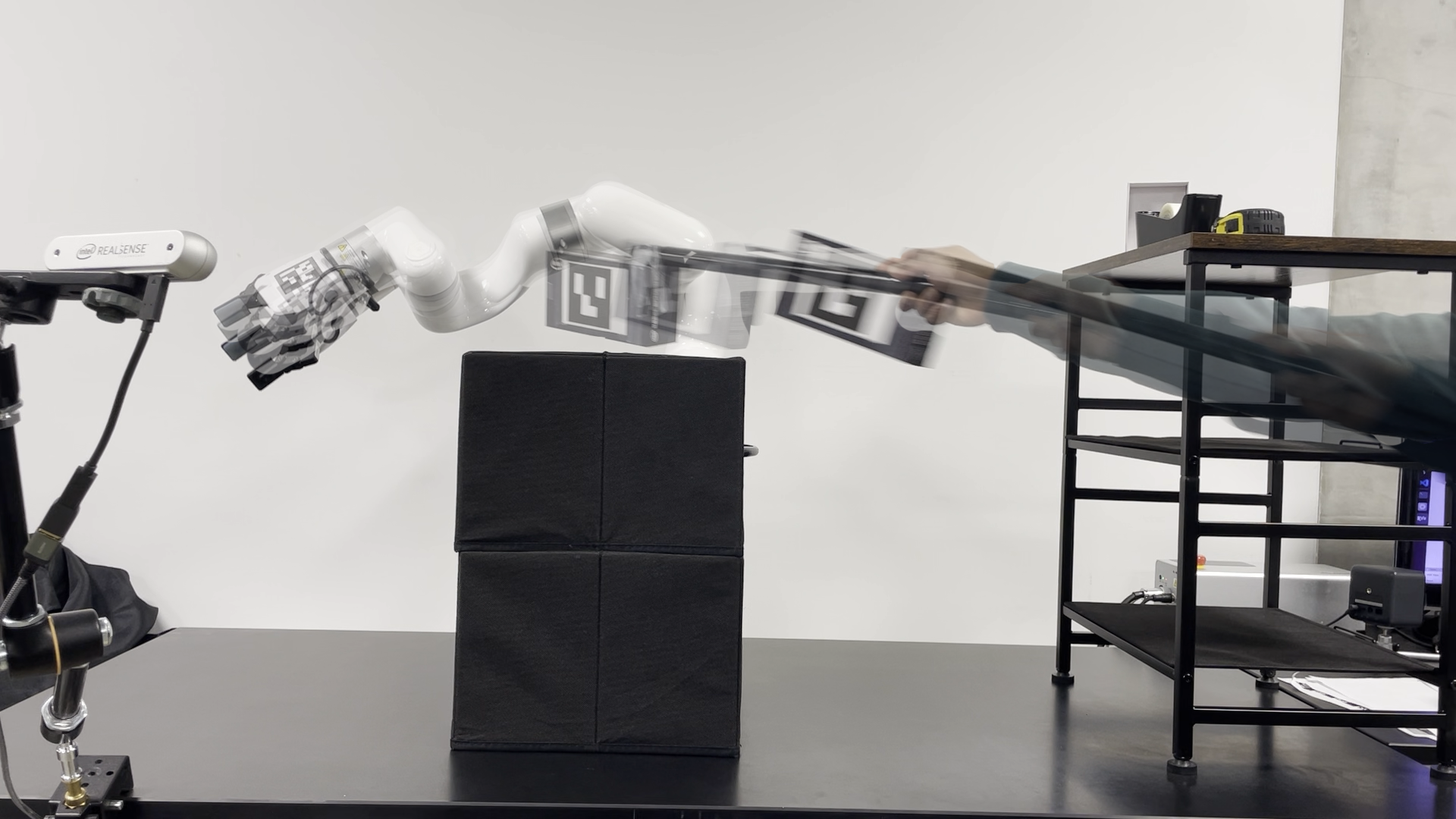}}
    \subcaptionbox{Response to dynamic obstacle\label{fig:7c}}
    {\includegraphics[width=0.24\textwidth]{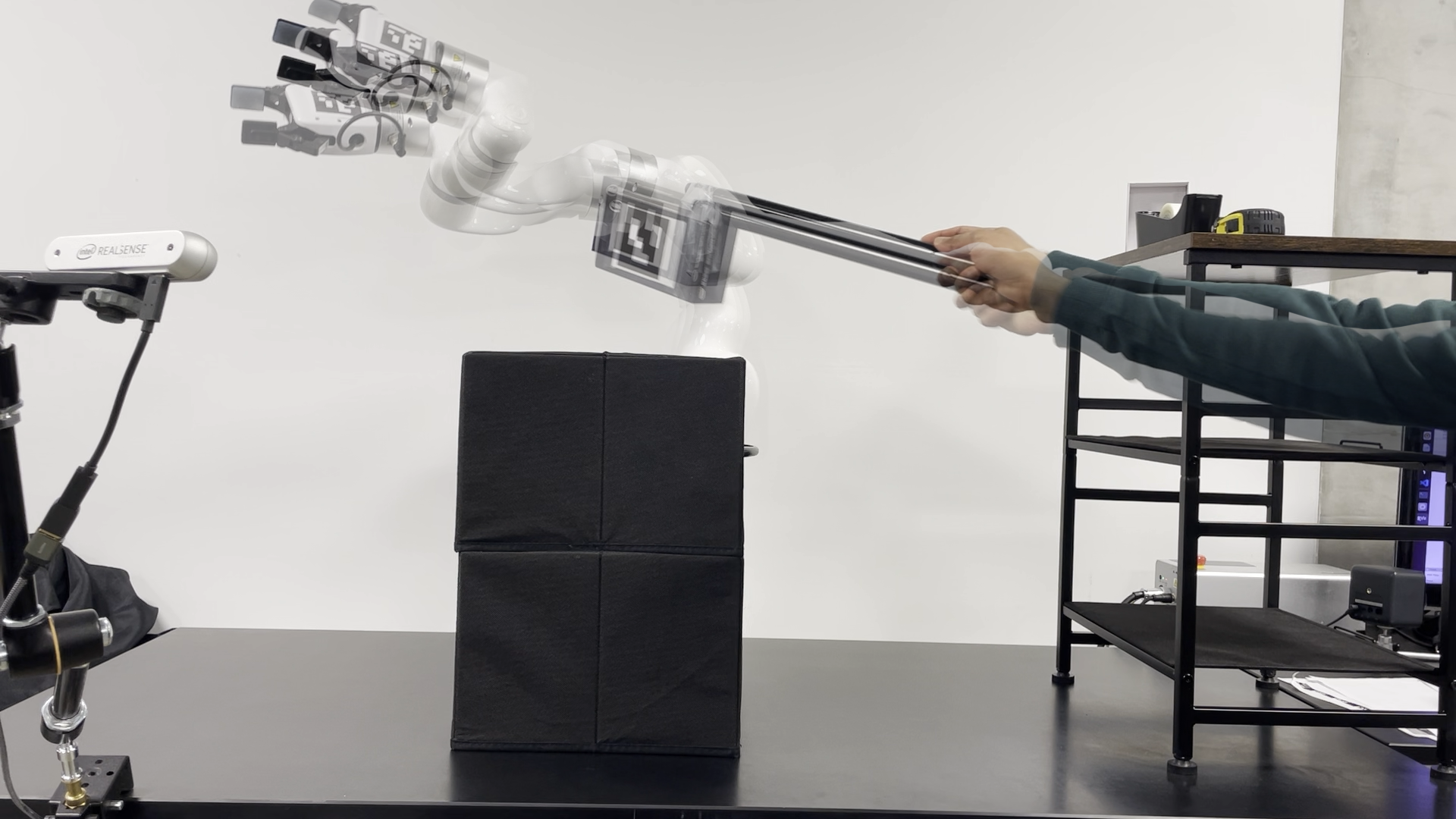}}
    \subcaptionbox{Response to dynamic obstacle\label{fig:7d}}
    {\includegraphics[width=0.24\textwidth]{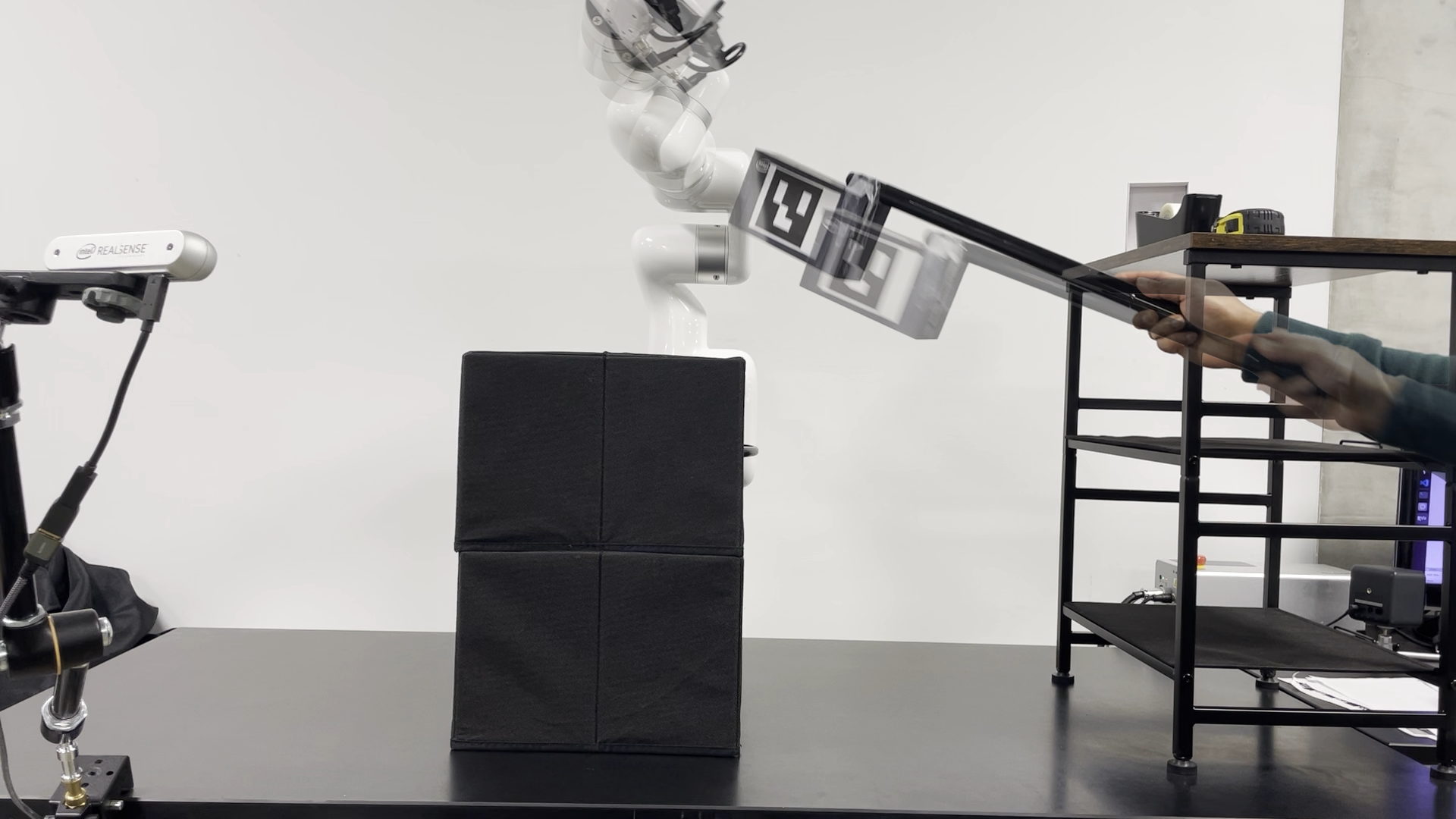}}  
    \caption{Bubble-CDF planner and DR-CBF control applied to two real-world setups for a 6-DoF xArm robot. 
    The top row (a-d) represents Setup 1, with the robot navigating a cluttered environment featuring a combination of static and dynamic obstacles. 
    Similarly, the bottom row (e-h) depicts Setup 2, showcasing the planner’s adaptability in a different obstacle layout. 
    For both setups: (a, e) illustrate the bubble-CDF planned configuration in static environments, while (b-d, f-h) demonstrate the robot's real-time adaptive responses to dynamic obstacles.}
    \label{fig: real_world_demo}
    \vspace*{-3ex}
\end{figure*}

We validate the efficiency of the bubble-CDF planner and the robustness of the DR-CBF controller in real-world experiments conducted on an xArm6 manipulator operating in cluttered and dynamic environments.

\textbf{Setup:} A 6-DoF xArm is mounted on a table within a workspace containing multiple static obstacles (Fig.~\ref{fig: real_world_demo}). A depth camera provides point-cloud observations of the scene. We perform 20 randomized trials by varying obstacle configurations and selecting different goal poses. Motion planning is conducted assuming a static environment. During execution, the robot tracks the planned trajectory while reacting to dynamic obstacles. The velocities of the moving objects are estimated in real time using ArUco markers.

\textbf{Results and Discussion:} Table~\ref{tab: real_xarm_planning_results} reports planning performance across all evaluated methods. We report the end-to-end runtime (including both planning and post-processing time) for all methods in this real-robot experiment.
The bubble-CDF planner achieves a substantial reduction in collision checks (nearly an order of magnitude fewer than baseline approaches) due to its use of configuration-space bubbles that certify local collision-free regions and reduce reliance on edge-based collision checking. Path lengths remain comparable across all planners, indicating that efficiency gains do not compromise solution quality, consistent with our simulation findings.

Fig.~\ref{fig: real_world_demo} shows snapshots of the bubble-CDF planner and DR-CBF controller in two different workspace arrangements.  The robot exhibits similar behavior to the simulation, following the nominal planned trajectory (Figs.~\ref{fig:6a},~\ref{fig:7a}) but avoiding dynamic obstacles as they are presented. Video recordings of these experiments are shown in the project webpage \url{https://existentialrobotics.org/bubble_cdf_safe_manipulation/}.

\section{Conclusion}
\label{sec:conclusion}

In this work, we proposed an integrated framework for safe motion planning and control of robotic manipulators in dynamic and cluttered environments. Our approach combines a novel bubble-CDF planner for efficient trajectory generation and a distributionally robust control barrier function (DR-CBF) formulation for safe navigation. Comprehensive evaluations, conducted in both simulation and real-world settings, validate the effectiveness of our approach, demonstrating high success rates and significant reductions in computational overhead. 

The bubble-CDF planner leverages the neural configuration-space distance function (CDF) to construct local safety regions, termed \textit{configuration-space bubbles}, enabling efficient exploration of the configuration space. By certifying entire regions as collision-free with a single query, the planner significantly reduces the number of collision checks compared to conventional sampling-based methods. This efficiency is achieved without compromising the quality of the planned trajectories. Additionally, the planner allows trajectory optimization to be formulated as a convex program, enabling smooth and safe trajectory refinement without requiring additional collision checks.

For real-time trajectory execution, we introduced a DR-CBF-QP safety filter that ensures the robot remains within a safe set throughout its motion. This formulation explicitly incorporates uncertainties in the neural CDF model, gradient estimates, and sensor measurements, leveraging a distributionally robust optimization framework. The DR-CBF-QP controller dynamically adapts to obstacles and sensor inaccuracies, maintaining safety while ensuring smooth trajectory tracking in both static and dynamic environments. An important direction for future work is the development of explicit feasibility recovery strategies for the DR-CBF-QP when uncertainty sets, control limits, or environment dynamics render the optimization infeasible.


\bibliography{references}

\end{document}